\documentclass[11pt]{article}

\usepackage{acl}
\usepackage{times}
\usepackage{latexsym}
\usepackage[T1]{fontenc}
\usepackage[utf8]{inputenc}
\usepackage{microtype}
\usepackage{inconsolata}
\usepackage{graphicx}
\usepackage{booktabs}
\usepackage{multirow}
\usepackage{array}
\usepackage{enumitem}
\usepackage{xcolor}
\usepackage{amsmath}
\usepackage{makecell}
\usepackage{tikz}
\usepackage{cuted}
\usepackage{placeins}
\usepackage{flafter}
\usepackage{multicol}
\usepackage{tabularx}
\usepackage{capt-of}
\usepackage{needspace}
\usepackage{float}
\usepackage{ragged2e}
\usepackage{balance}
\usepackage{comment}
\usepackage{xurl}
\usepackage{listings}
\usepackage{xcolor}

\lstdefinestyle{prompt}{
   basicstyle=\ttfamily\scriptsize,
   columns=fullflexible,
   keepspaces=true,
   showstringspaces=false,
   breaklines=true,
   breakatwhitespace=true,
   breakindent=0pt,
   breakautoindent=false,
   postbreak={},
   frame=single,
   rulecolor=\color{black!40},
   framexleftmargin=0pt,
   aboveskip=0.5\baselineskip,
   belowskip=0.5\baselineskip
}
\usetikzlibrary{arrows.meta,positioning,shapes.geometric,calc}

\definecolor{pipelineData}{RGB}{226,238,247}
\definecolor{pipelineDataBorder}{RGB}{72,112,138}
\definecolor{pipelineLLM}{RGB}{239,231,244}
\definecolor{pipelineLLMBorder}{RGB}{112,82,126}
\definecolor{pipelineMT}{RGB}{250,235,216}
\definecolor{pipelineMTBorder}{RGB}{155,106,55}
\definecolor{pipelineLoc}{RGB}{226,241,232}
\definecolor{pipelineLocBorder}{RGB}{69,125,91}

\newcommand{\midsize}{\fontsize{9.5pt}{11pt}\selectfont}

\newcommand{\middata}{\textsc{\midsize Task-Preserved}}
\newcommand{\loc}{\textsc{EuroAlpaca}}
\newcommand{\dataset}{\textsc{\midsize EuroAlpaca}}
\newcommand{\datasettitle}{\textls[-15]{\textsc{EuroAlpaca}}}
\newcommand{\benchmark}{\textsc{\midsize European-IFEval}}
\newcommand{\benchmarktitle}{\textsc{European-IFEval}}
\newcommand{\benchmarktable}{\textls[-20]{\textsc{European-IFEval}}}
\newcommand{\datasettable}{\textls[-50]{\textsc{EuroAlpaca}}}
\newcommand{\datasettablee}{\textls[-20]{\textsc{EuroAlpaca}}}

\title{\datasettitle{}: Task-Preserving Localisation of Instruction Data \\ for European Languages}

\author{
  Aleix Sant$^{1,2}$\textbf{, }
  Jordi Luque$^{1}$\textbf{, }
  Carlos Escolano$^{2}$ \\[4pt]
  $^{1}$Scientific Research, Telefónica Innovación Digital \\
  $^{2}$Universitat Politècnica de Catalunya \\
  Barcelona, Spain \\[2pt]
  \texttt{\small \{aleix.santsavall,jordi.luque\}@telefonica.com} \\
  \texttt{\small carlos.escolano@upc.edu}
}

\begin{document}
\maketitle

\begin{abstract}
Machine translation (MT) offers a scalable way to extend English instruction-tuning data to multiple languages, but it can distort task-critical constraints and required outputs, creating corrupted training examples and degrading models trained on such data. We introduce \dataset{}, a task-preserving localisation pipeline and near-parallel resource covering 50 European languages and regional varieties, together with \benchmark{}, a multilingual benchmark for verifiable instruction following. Depending on the example, our pipeline applies field-wise MT while preserving task-critical content or reconstructs a task-equivalent target-language instance, followed by validation of cross-field coherence and target-language consistency. Across LoRA experiments with four LLMs, training on directly translated data improves ROUGE-L and F\textsc{bert} on the Aya Evaluation Suite, but reduces accuracy on \benchmark{} by 29.8\% relative to the unadapted baseline. In contrast, adaptation with \dataset{} improves accuracy by 12.9\% over the same baseline, reversing the degradation caused by direct MT, %recovering the loss from direct MT and yielding a net improvement, 
while also achieving the highest ROUGE-L and F\textsc{bert} scores on Aya. These results show that preserving task semantics is essential for multilingual instruction tuning.
\end{abstract}

\section{Introduction}
Instruction tuning enables language models to follow diverse natural-language requests and generalise to unseen tasks \citep{DBLP:conf/iclr/WeiBZGYLDDL22, DBLP:conf/nips/Ouyang0JAWMZASR22}. However, open instruction-tuning resources remain predominantly English-centric. These corpora encompass heterogeneous tasks, including question answering, summarisation, classification, reasoning, rewriting, constrained generation and code-related tasks. A common scalable strategy for extending them to new languages is to machine-translate English instruction datasets, including the textual fields of each instruction--input--output
record \citep{DBLP:journals/corr/abs-2311-10797, DBLP:conf/acl/SinghVD0MKSPMOZ24, DBLP:conf/eacl/ChenJBKHH24, DBLP:conf/acl/0002HASTE24}. Although scalable, this strategy conflates preserving sentence meaning with preserving the task encoded by the complete example~\citep{DBLP:journals/corr/abs-2107-04512}.

The distinction matters when a task depends on linguistic form or on the role of a particular field. Previous work has shown that translated instruction data can corrupt language-dependent tasks such as grammatical-error correction, phonological constraints and examples containing code or language-specific knowledge \citep{DBLP:conf/emnlp/ChenYGH24, DBLP:conf/acl/LiYZLWZ24}. For example, translating an intentionally ungrammatical sentence may silently correct the error that the model is expected to identify. Similarly, constraints involving word length, rhyme, alliteration or acronym formation are not generally invariant under MT, because they depend on the linguistic form of the target-language expression. Translating classification labels, source code, URLs, paths or required output strings can likewise make an otherwise fluent example inconsistent. In such cases, we use the term \emph{task-preserving localisation} to mean constructing a target-language instruction–input–output triple that preserves the underlying task, its explicit constraints, its difficulty and the answer relation, even when its surface form is not a literal translation.

In line with~\citealp{DBLP:conf/eamt/SkadinsPVSH14}, we argue that task preservation is a central requirement for synthetic instruction data adaptation in new and low-resource languages. Fluency alone is insufficient. Translated supervision may appear natural while weakening a model's ability to satisfy verifiable instructions. We therefore formulate multilingual instruction-data generation as a structured decision process rather than uniform machine translation. Each example is treated as a structured instruction-following instance, and LLM-based judge modules determine whether it is suitable for direct translation, which fields must be preserved, and when task-level rewriting is required. % We therefore formulate localization as a structured decision process, where text generation is only one stage of preserving the original supervision signal.

Starting from the 51,760 examples in Alpaca Cleaned\footnote{\url{https://github.com/gururise/AlpacaDataCleaned}}, we build \dataset{}. Our proposed data processing pipeline (i) annotates the task and domain, (ii) decides whether the sample can be machine-translated, (iii) judges which fields must remain unchanged, (iv) translates or rewrites the example for a target language $\ell$, and (v) validates the complete target-language triple for coherence and language consistency. The resulting resource is near-parallel, i.e. examples remain aligned by source identifier and task intent, but language-dependent content may differ when literal equivalence would break the task. This makes task-preserving localisation a constrained NLG problem rather than a purely translation-based one, since some target-language content must be adapted or reconstructed to preserve the task.
%Note that we frame task-preserving localization as a constraint-based NLG task rather than passive machine translation, synthesizing new target-language content to satisfy logical and formal rules.

We evaluate the usefulness of \dataset{}, our localised version of the Alpaca Cleaned instruction dataset, by language-specific fine-tuning across four contemporary, state-of-the-art small-sized language models. By leveraging the Aya Evaluation Suite alongside our newly curated \benchmark{} benchmark, we provide a comprehensive analysis of 50 European languages, which reveals an evaluation mismatch. Direct MT improves reference similarity but substantially reduces instruction-following accuracy. In contrast, models trained with \dataset{} consistently improve both, supporting our assumption that multilingual instruction data should preserve task function rather than only target-language fluency and sentence meaning. Our contributions are:
\begin{itemize}[leftmargin=*,nosep]
    \item An LLM-guided pipeline for task-preserving multilingual instruction-data generation.
    \item \dataset{}, a source-aligned, near-parallel resource covering English and 50 European languages and regional varieties.
    \item \benchmark{}, a normalised multilingual instruction-following benchmark collection with explicit source provenance across the same language set.
    \item Controlled language-specific experiments across four instruction-tuned models.
    \item Matched comparisons with multilingual instruction datasets, including Bactrian-X, Okapi and MITS.
\end{itemize}
\section{Related Work}

\paragraph{Synthetic instruction data.}
Self-Instruct and Alpaca showed that model-generated instructions can cheaply expand supervised instruction-tuning data \citep{DBLP:conf/acl/WangKMLSKH23, taori2023alpaca}. Alpaca Cleaned corrected flawed records from the original release. These resources follow a simple schema consisting of an \emph{instruction}, which defines the task to be performed, an optional \emph{input}, which provides additional context for the task, and an \emph{output}, which contains the reference response expected from the model.

\paragraph{Multilingual instruction-tuning resources.}
\label{resources}
Several efforts extend English instruction data to multiple languages through translation, templating, human curation and preference annotation. Okapi translates Alpaca-52K into 26 languages using ChatGPT and provides multilingual response-ranking data for RLHF \citep{DBLP:conf/emnlp/LaiNNNDRN23}. MITS translates Alpaca-52K-GPT4 and Dolly-15K into 132 languages with Google Cloud Translation \citep{DBLP:journals/corr/abs-2311-10797}. Bactrian-X translates Alpaca- and Dolly-style prompts and pairs them with target-language responses generated by \texttt{gpt-3.5-turbo} \citep{DBLP:journals/corr/abs-2305-15011}. The Aya Collection combines multilingual instruction--response data from multiple sources and construction strategies \citep{DBLP:conf/acl/SinghVD0MKSPMOZ24}.

\paragraph{Instruction evaluation beyond English.}
%\paragraph{Multilingual instruction-following evaluation.}
IFEval evaluates compliance with objectively verifiable instructions rather
than relying solely on reference-based answer similarity \citep{DBLP:journals/corr/abs-2311-07911}. Multilingual extensions include M-IFEval, which covers French, Japanese and Spanish, and Marco-Bench-MIF, a localised benchmark covering 30 languages \citep{DBLP:conf/naacl/DussolleDSD25, DBLP:conf/acl/ZengLLZWNSZLZLG25}, and XIFBench~\citep{DBLP:conf/nips/LiCLBZWLZ25}, introducing fine-grained multilingual constraint analysis. The Aya Evaluation Suite complements constraint-based evaluation with multilingual open-ended generation tasks and reference-based metrics \citep{DBLP:conf/acl/SinghVD0MKSPMOZ24}. LLM-as-a-judge approaches provide an alternative for open-ended evaluation, although their judgements may exhibit evaluator and presentation biases \citep{DBLP:conf/emnlp/LiuIXWXZ23, DBLP:conf/nips/ZhengC00WZL0LXZ23}.

\paragraph{From direct translation to task localisation.}
Direct translation provides a scalable way to construct multilingual instruction data, but it can introduce translationese, create knowledge--language mismatches and corrupt language-dependent tasks

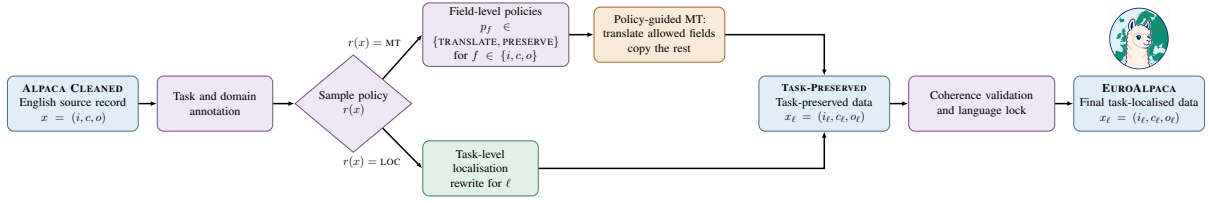
\begin{figure*}[!t]
\centering
\resizebox{0.99\textwidth}{!}{%
\begin{tikzpicture}[
    font=\normalsize,
    node distance=5mm and 6mm,
    line width=0.7pt,
    arr/.style={
        -{Latex[length=2mm,width=1.6mm]},
        line width=1.4pt
    },
    data/.style={
        draw=pipelineDataBorder,
        fill=pipelineData,
        rounded corners=6pt,
        align=center,
        minimum height=18mm,
        text width=40mm,
        inner sep=4pt
    },
    llm/.style={
        draw=pipelineLLMBorder,
        fill=pipelineLLM,
        rounded corners=6pt,
        align=center,
        minimum height=18mm,
        text width=45mm,
        inner sep=4pt
    },
    llm_small/.style={
        draw=pipelineLLMBorder,
        fill=pipelineLLM,
        rounded corners=6pt,
        align=center,
        minimum height=18mm,
        text width=35mm,
        inner sep=4pt
    },
    decision/.style={
        draw=pipelineLLMBorder,
        fill=pipelineLLM,
        diamond,
        aspect=1.2,
        align=center,
        text width=25mm,
        inner sep=2.1pt
    },
    mt/.style={
        draw=pipelineMTBorder,
        fill=pipelineMT,
        rounded corners=6pt,
        align=center,
        minimum height=18mm,
        text width=40mm,
        inner sep=4pt
    },
    localisation/.style={
        draw=pipelineLocBorder,
        fill=pipelineLoc,
        rounded corners=6pt,
        align=center,
        minimum height=18mm,
        text width=35mm,
        inner sep=4pt
    },
    edge-label/.style={
        font=\normalsize,
        fill=white,
        inner sep=1pt
    }
]

% Source and annotation
\node[data] (src)
{\textsc{\textbf{Alpaca Cleaned}}\\
English source record\\
$x=(i,c,o)$};

\node[llm_small, right=6mm of src] (ann)
{Task and domain\\
annotation};

% Sample-level routing
\node[decision, right=7mm of ann] (gate)
{Sample policy\\
$r(x)$};

% Field-level routing and MT branch
\node[llm, above right=4mm and 13mm of gate] (policy)
{Field-level policies\\
$p_f\in\{\textsc{translate},\textsc{preserve}\}$\\
for $f\in\{i,c,o\}$};

\node[mt, right=8mm of policy] (mtbox)
{Policy-guided MT: \\
translate allowed fields \\
copy the rest};

% Task-localisation branch
\node[localisation, below right=4mm and 13mm of gate] (locbox)
{Task-level\\
localisation\\
rewrite for $\ell$};

% Merge and validation
\node[data, right=203mm of src] (mid)
{\textbf{\middata{}}\\
Task-preserved data\\
$x_\ell=(i_\ell,c_\ell,o_\ell)$};

\node[llm, right=6mm of mid] (val)
{Coherence validation\\
and language lock};

\node[data, right=6mm of val] (out)
{\textbf{\loc{}}\\
Final task-localised data\\
$x_\ell=(i_\ell,c_\ell,o_\ell)$};

\node[
    anchor=south,
    inner sep=0pt,
    yshift=2mm
] at (out.north) {
    \includegraphics[width=20mm]{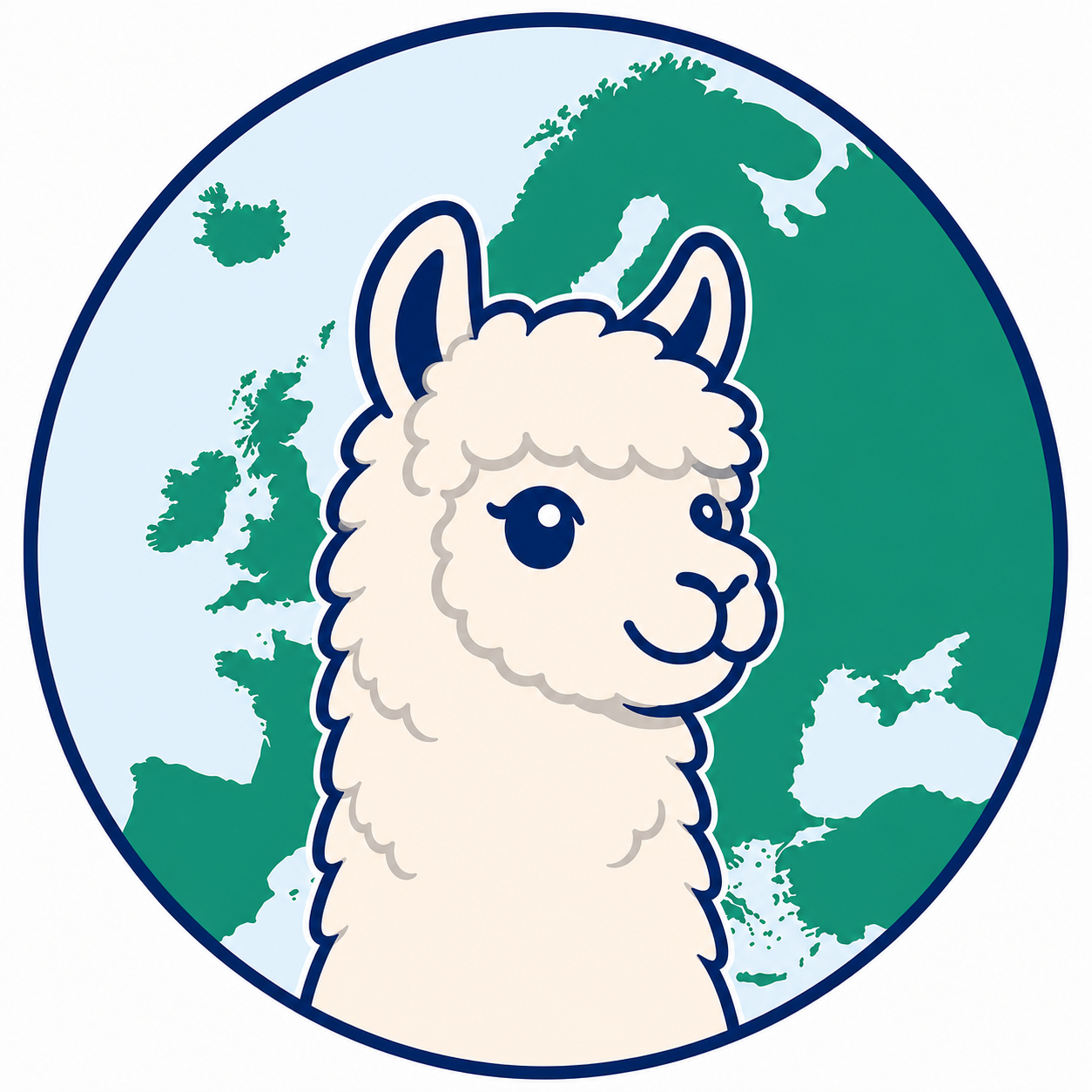}
};

% Main flow
\draw[arr] (src) -- (ann);
\draw[arr] (ann) -- (gate);

% Sample passes MT gate
\draw[arr]
    (gate.north east)
    -- node[edge-label, above left, yshift=2mm]
        {$r(x)=\textsc{mt}$}
    (policy.west);

\draw[arr] (policy) -- (mtbox);

\draw[arr]
    (mtbox.east)
    -| (mid.north);

% Sample requires task-level localisation
\draw[arr]
    (gate.south east)
    -- node[edge-label, below left, yshift=-2mm]
        {$r(x)=\textsc{loc}$}
    (locbox.west);

\draw[arr]
    (locbox.east)
    -| (mid.south);

% Final validation
\draw[arr] (mid) -- (val);
\draw[arr] (val) -- (out);

\end{tikzpicture}%
}
\caption{Task-preserving instruction-data generation pipeline. Blue nodes are data artefacts, purple are LLM stages, orange is policy-guided MT, and green is task-level rewriting. Gated routing directs samples to either per-field translation (\textsc{translate}/\textsc{preserve}) or task-level localisation, yielding \middata{} which is validated to produce \loc{} for language $\ell$.}

\label{fig:pipeline}
\end{figure*}

\noindent
such as grammatical error correction~\citep{DBLP:conf/emnlp/ArtetxeLA20, DBLP:conf/acl/RileyCFG20, DBLP:conf/emnlp/ChenYGH24, DBLP:journals/corr/abs-2504-10356}. Standard supervised fine-tuning (SFT) may further amplify translationese biases present in translated training data~\citep{DBLP:conf/acl/LiZWZCYXZ25}. Training-side methods such as X-CIT improve learning from parallel translated instructions but do not directly address whether translation preserves the validity of the underlying task~\citep{DBLP:conf/acl/WuWY025}. Other approaches move beyond direct translation by deriving instruction--response pairs from human-written target-language texts or by localising communicative intent and language- or culture-dependent content~\citep{DBLP:journals/tacl/KoksalTIUKS25, DBLP:conf/emnlp/VasselliKSW25, DBLP:conf/naacl/MaheshwaryYNMM25}. Building on this shift towards localisation, our pipeline instead targets task-preserving localisation of \emph{instruction}–\emph{input}–\emph{output} triples.
\section{Task-Preserving Localisation}
\label{sec:task-localisation}

Let a source record be $x=(i,c,o)$, containing an instruction $i$, an
optional input $c$ and an output $o$. For a target language $\ell$, the goal
is to construct $x_\ell$ such that the task relation $T(i,c,o)$ remains valid. Literal semantic equivalence of every field is neither necessary nor sufficient.

We use a hierarchical routing process. First, a sample-level policy $r(x)\in\{\textsc{mt},\textsc{loc}\}$ determines whether the record is suitable for the field-wise MT route or instead requires task-level localisation. Only samples routed to \textsc{mt} undergo a second pass, which assigns each non-empty field $f\in\{i,c,o\}$ a policy $p_f\in\{\textsc{translate},\textsc{preserve}\}$. Samples routed to \textsc{loc} bypass field-wise MT and are reconstructed as task-equivalent target-language instances.

\subsection{When Translation Breaks the Task}

Table~\ref{tab:failure-taxonomy} shows the operational taxonomy used by the sample policy prediction. It distinguishes cases 

\begin{table}[!t]
\centering
\small
\begin{tabular}{lrr}
\toprule
\textbf{Handling regime} & \textbf{Count} & \textbf{\%} \\
\midrule
Direct translation & 27,946 & 54.0 \\
Field-preserving translation & 12,726 & 24.6 \\
Task-level localisation & 11,088 & 21.4 \\
\midrule
Total & 51,760 & 100.0 \\
\bottomrule
\end{tabular}
\caption{Handling regimes for the 51,760 source examples. Direct translation translates all non-empty fields. Field-preserving translation retains at least one field unchanged. Task-level localisation reconstructs examples for which field-wise MT would invalidate the task.
}
\label{tab:handling-regimes}
\end{table}

\noindent
that require target-language rewriting from cases in which specific fields should remain unchanged.

Table~\ref{tab:schematic-examples} provides schematic illustrations of these examples, while Tables~\ref{tab:appendix-localisation-examples} and~\ref{tab:appendix-preservation-examples} in the Appendix present concrete examples of the different possible handling regimes. The appropriate handling depends on where task-critical information is encoded across the \emph{instruction}, \emph{input} and \emph{output} fields.

\subsection{Task-Preserving Data Generation Pipeline}
Figure~\ref{fig:pipeline} illustrates the complete pipeline. We use an 8-bit quantised variant of \texttt{google/gemma-4-31B-it}~\citep{google2026gemma4} as the LLM-based decision module, with specialised prompts for semantic annotation, sample- and field-level policy predictions, task-preserving localisation (i.e., rewriting in the target language) and consistency validation with target-language enforcement. Complete inference settings and prompt templates for the different steps are reported in Appendix~\ref{app:reproducibility}. Prompt templates were iteratively refined through manual inspection of sampled outputs. LLM outputs are parsed as JSON and normalised against stage-specific formats defined in the prompts. Malformed outputs are handled conservatively. Sample policy prediction failures fall back to non-translatable decisions, while task-level localisation and consistency validation retry with an explicit repair prompt.

\begin{table*}[!t]
\centering
\small
\begin{tabular}{p{0.21\textwidth}p{0.33\textwidth}p{0.34\textwidth}}
\toprule
\textbf{Failure type} & \textbf{Typical cases} & \textbf{Handling policy} \\
\midrule
Source-language form dependence
& Grammar or spelling correction, punctuation judgements, tense/voice edits, alphabetic ordering
& Generate an analogous target-language task and recompute the output. \\
\midrule
Lexical or phonological constraints
& Letter/character counts, word counts, rhymes, acrostics, mnemonics, puns, anagrams
& Replace the constrained material with target-language content satisfying the same constraint. \\
\midrule
Language- or culture-specific expressions
& Idioms, proverbs, fixed expressions, English-only prompts, language-specific examples
& Substitute a natural target-language analogue while preserving task intent and difficulty. \\
\midrule
Protected literals
& Code, URLs, paths, identifiers, formulas, quoted strings, names, titles, table columns
& Preserve fields containing task-critical literal artefacts. \\
\midrule
Fixed output vocabulary or schema
& Classification labels, JSON keys, structured outputs, canonical options
& Preserve fields defining or containing the required labels or output format. \\
\midrule
Field-role dependence
& Translation tasks, demonstrations, input--output examples, source/target language roles
& Preserve the role of each field rather than translating all fields uniformly. \\
\bottomrule
\end{tabular}
\caption{Operational taxonomy of examples requiring non-uniform translation across fields.}
\label{tab:failure-taxonomy}
\end{table*}

\begin{table*}[!t]
\centering
\small
\begin{tabular}{p{0.22\textwidth}p{0.32\textwidth}p{0.34\textwidth}}
\toprule
\textbf{Source task} & \textbf{Uniform MT failure} & \textbf{Task-preserving construction} \\
\midrule
Correct the grammar in an intentionally erroneous sentence.
& Translation may silently remove the error, leaving nothing to correct.
& Create a comparable target-language error and provide its correction. \\
\midrule
Return a word that rhymes with a given word.
& Rhyme depends on target-language phonology. Translated words may not rhyme.
& Select a target-language cue and answer satisfying the same relation. \\
\midrule
Write exactly ten words about a topic.
& Word counts are not preserved under translation.
& Generate target-language text satisfying the count. \\
\midrule
Classify with exact labels.
& Translating labels can break the required output vocabulary or automatic checker.
& Translate the content but preserve the labels verbatim. \\
\midrule
Interpret an idiom.
& Literal translation may produce a nonsensical expression or a different idiom.
& Replace it with a natural target-language idiom and explain that expression. \\
\midrule
Translate a sentence from language A to language B.
& Translating every field can invalidate the specified language pair.
& Translate the instruction, but preserve the source and target texts. \\
\midrule
Write a function in a specified programming language for a given operation.
& Translating code can alter keywords, identifiers, literals or syntax.
& Translate the natural-language instruction, but preserve the generated code exactly. \\
\bottomrule
\end{tabular}
\caption{Schematic examples where direct field-wise translation can break the target-language instruction task.}
\label{tab:schematic-examples}
\end{table*}

\paragraph{Semantic annotation.}
For each source example, we prompt Gemma-4 to assign one primary task and primary domain, as well as optional secondary tasks and domains, using predefined taxonomies (Appendix~\ref{distributions}). These annotations support corpus analysis and inform later sample- and field-level translation-policy predictions. In these later stages, they are provided as context to Gemma-4 and used by deterministic guardrails. The pipeline ensures that all predicted labels belong to the predefined taxonomies. For example, it replaces missing or invalid primary labels with \emph{other} for tasks and \emph{general} for domains. %The resulting primary task label is also used by deterministic guardrails in later routing decisions. 
The resulting distributions reveal substantial imbalances across tasks and domains. %canviar referencia dalt i ficarala aqui quan creei taxonomies

\paragraph{Sample- and field-level policy predictions.}
The sample-level gate first decides whether a record is suitable for field-wise MT or instead requires task-level localisation through rewriting. This gate combines a global include/exclude prediction from Gemma-4 with deterministic guardrails that can override the model's prediction and reject samples whose task depends on source-language form, such as spelling/grammar exercises, rhyme, mnemonics, idioms... For samples that pass the gate, Gemma-4 predicts a translation policy for each field: \emph{instruction}, \emph{input} and \emph{output}. Deterministic field-level guardrails then adjust these predictions when necessary. Table~\ref{tab:handling-regimes} reports the resulting regimes.

\paragraph{Policy-guided MT and task localisation.}
For MT-routed samples, NLLB-3.3B \citep{DBLP:journals/corr/abs-2207-04672} translates fields assigned \textsc{translate}, while fields assigned \textsc{preserve} are copied unchanged. Samples routed to localisation are instead reconstructed by Gemma-4 using a curated prompt, preserving the task type, difficulty, expected answer form and explicit constraints in the target language. These two routes then produce the intermediate \middata{} artefact, which preserves task function. Consequently, some MT-routed samples may retain English content in preserved fields when translating it would invalidate the task.

\paragraph{Consistency validation and target-language enforcement.}
In the final step, Gemma-4 examines the complete \emph{instruction}--\emph{input}--\emph{output} triple for semantic contradictions, unclear wording, grammatical or orthographic errors, literal calques, mixed-language content and instruction--response mismatches. It enforces target-language coherence, applies recoverable edits and records post-edit metadata, while preserving non-target-language spans if a specific task requires it.

\section{Generated Resources}
\label{sec:resources}

\subsection{Near-Parallel Instruction Data}
We run the pipeline for 50 European languages and regional varieties, in addition to the annotated English source\footnote{\url{https://github.com/Telefonica-Scientific-Research/EuroAlpaca}}. Each target record retains the source identifier, language tag, semantic labels, whole-sample and field policies, validation outcome and post-edit flags. The collection is \emph{near-parallel}, with records aligned by source identifier and task intent while allowing lexical or structural differences when literal equivalence would invalidate the task.

\begin{figure}[!t]
\centering
\includegraphics[width=\columnwidth]{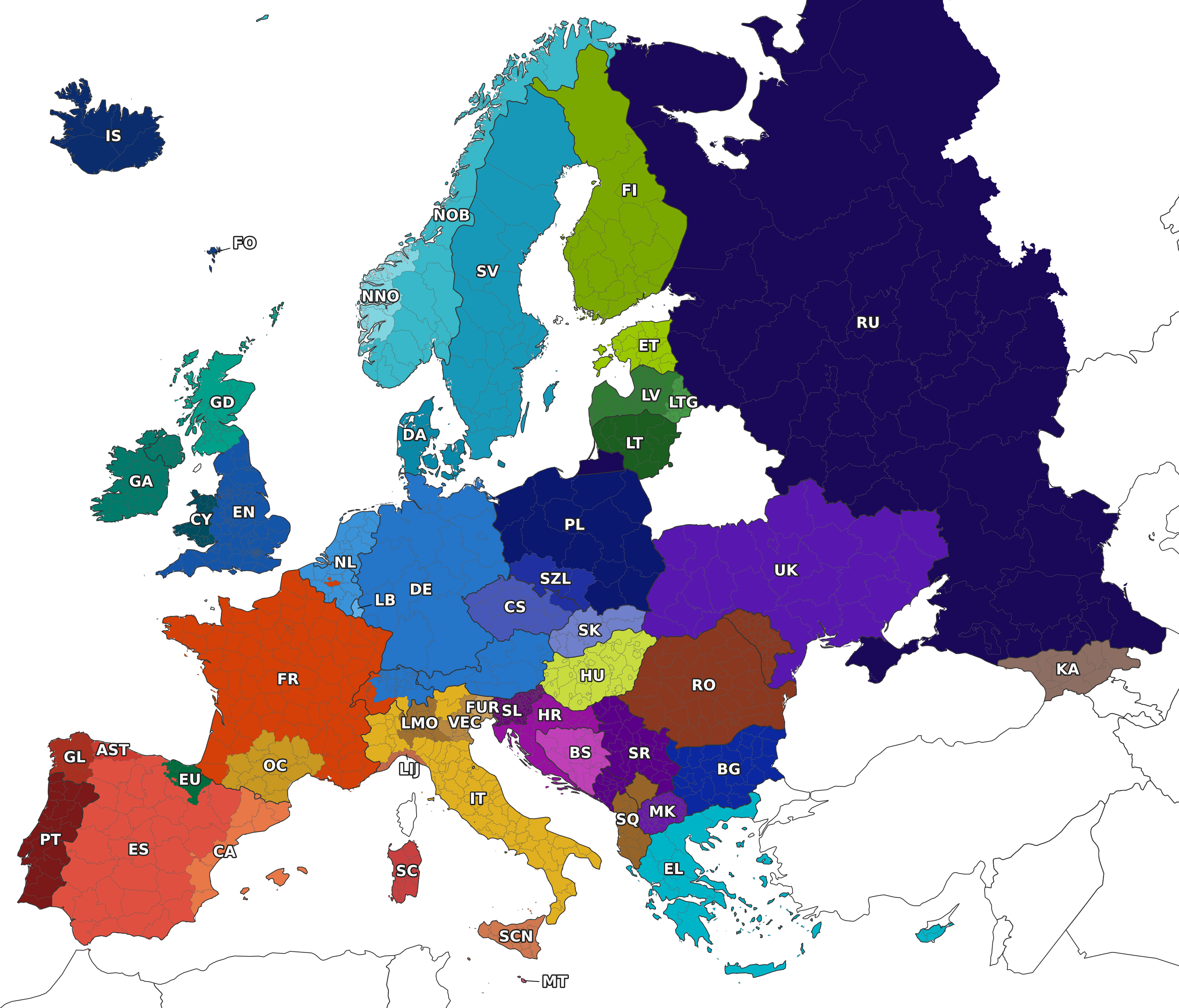}
\caption{Approximate traditional geographic areas associated with the 50 European languages and regional varieties covered by \dataset{} and \benchmark{}. Boundaries are schematic and do not represent exclusive or current speaker distributions. The map is based on \citealp{fuen2013language}.}
\label{fig:map}
\end{figure}

The language set is European-centric but typologically diverse, as shown in Figure~\ref{fig:map} and Appendix Table~\ref{tab:language-coverage}. It includes major national languages as well as regional and minority varieties from several language families and branches, including Romance, Germanic, Slavic, Celtic, Baltic, Uralic, Hellenic, Albanian, Semitic and Kartvelian, together with language isolates. It also spans the Latin, Cyrillic, Greek and Georgian scripts. We define resource tiers using clean monolingual token counts from MADLAD-400 \citep{DBLP:conf/nips/KuduguntaC0GXKS23}, with high denoting $\geq 5$ billion tokens, medi-

\noindent
um denoting $100$ million to $<5$ billion, and low denoting $<100$ million tokens or no separate clean monolingual partition in MADLAD-400. Combined with the near-parallel design of \dataset{}, this diversity enables controlled multilingual instruction-following experiments across languages that differ in typology, script and resource availability, while keeping the underlying supervision aligned.

\subsection{\benchmarktitle{}}
\label{sec:european-ifeval}

\benchmark{} is a separate evaluation resource, not an output of the \dataset{} localisation pipeline\footnote{\url{https://github.com/Telefonica-Scientific-Research/european-ifeval}}. It brings together existing IFEval-style resources for the same 50 target languages and converts them to a common evaluation format. For each language, we select one source using a fixed quality-oriented hierarchy. We first prioritise native-speaker-curated or professionally translated language-specific releases. When these are unavailable, we use Marco-Bench-MIF, which localises linguistic constraints and cultural references through automated processing followed by two rounds of human verification \citep{DBLP:conf/acl/ZengLLZWNSZLZLG25}. For the remaining languages, we use MultiIFEval, which translates and localises English IFEval prompts with an LLM while providing target-language Wikipedia content as contextual support.\footnote{
\url{https://huggingface.co/datasets/danish-foundation-models/multi-ifeval}}

We convert the gathered datasets to a common record format containing the prompt, instruction-checker identifiers and their arguments, a sample identifier and an indicator of whether the response is required to be in the target language. A separate manifest records, for each language, the source, provenance category, missing source items and information about explicit response-language constraints. We harmonise checker identifiers and argument formats so that all datasets can be scored with a common evaluator. Response-language constraints are evaluated using GlotLID-M, selected for its coverage of low-resource languages and language varieties \citep{DBLP:conf/emnlp/KargaranIYS23}.
\section{Experimental Setup}
\label{sec:experiments}

\subsection{Models and Supervision Conditions}
We evaluate four multilingual text-only instruction-tuned models in the 3--4B parameter range: Tiny-Aya-Global \citep{DBLP:journals/corr/abs-2603-11510}, Phi-4-Mini-instruct \citep{DBLP:journals/corr/abs-2503-01743}, Ministral-3-3B-Instruct-2512 \citep{mistral2025ministral3} and Llama-3.2-3B-Instruct \citep{meta2024llama32}.

For each of the 200 model–language pairs (four models and 50 languages), we train three separate language-specific LoRA adapters \citep{DBLP:conf/iclr/HuSWALWWC22}, one for each supervision condition described below, yielding 600 fine-tuning runs in total. Each adapter is trained on 50,760 examples and validated on the same 1,000 held-out examples for that language. We also evaluate the unadapted checkpoint as a baseline. The three supervision conditions contrast: (i) \textbf{Direct MT}, translating all fields uniformly with NLLB-3.3B; (ii) \textbf{Task-Preserved}, an ablation using the intermediate dataset resulting from sample-level routing, field-level preservation policies and task-level localisation; and (iii) \textbf{\datasettitle{}}, the final dataset produced after validation.

Within each model--language pair, all training and evaluation settings are
held fixed across supervision conditions, so performance differences primarily reflect the effect of the supervision data. We additionally fine-tune the same models on the Alpaca-derived subsets of Okapi, MITS and Bactrian-X, excluding non-Alpaca synthetic and preference data, and evaluate them using the same method. Each comparison is conducted on the languages shared by \dataset{}, the external resource and the corresponding benchmark. 

\subsection{Fine-Tuning}
All runs use LoRA with rank 16, alpha 32, dropout 0.05, and adapters on query, key, value and output projections.
We use AdamW, learning rate $10^{-5}$, bf16 mixed precision, gradient clipping at 1.0, micro-batches of 2 and 32 gradient-accumulation steps for an effective batch size of 64. The random seed is 42. Early stopping selects the final adapter.

\subsection{Evaluation and Aggregation}
We use two complementary evaluation settings to assess different aspects of multilingual instruction-tuned models. The Aya Evaluation Suite \citep{DBLP:conf/acl/SinghVD0MKSPMOZ24} evaluates open-ended language generation using ROUGE-L for lexical overlap \citep{lin2004rouge} and BERTScore for semantic similarity to reference responses \citep{DBLP:conf/iclr/ZhangKWWA20}. We report the F1 variant of BERTScore, denoted F\textsc{bert} throughout the paper.

In contrast, \benchmark{} measures compliance with verifiable instructions and reports strict and loose accuracy scores. In this work, we report only strict accuracy. Together, they distinguish reference similarity from instruction following. \benchmark{} covers all 50 languages, whereas Aya provides reference data for 36.

For each benchmark and metric, we macro-average scores across the available non-English languages within each model and then average the resulting means across the four models. Full per-language and per-model results are reported in Appendix~\ref{app:full-results}.

\begin{table*}[t]
\centering
\small
\setlength{\tabcolsep}{4.2pt}
\renewcommand{\arraystretch}{1.08}
\begin{tabular}{ll
                rr
                rr
                rr}
\toprule
& &
\multicolumn{2}{c}{\textbf{Aya ROUGE-L} $\uparrow$} &
\multicolumn{2}{c}{\textbf{Aya F\textsc{bert}} $\uparrow$} &
\multicolumn{2}{c}{\textbf{\benchmarktable{} Acc.} $\uparrow$} \\
\cmidrule(lr){3-4}
\cmidrule(lr){5-6}
\cmidrule(lr){7-8}
\textbf{Model} &
\textbf{Training data} &
\textbf{Mean} & \textbf{$\Delta$OTS} &
\textbf{Mean} & \textbf{$\Delta$OTS} &
\textbf{Mean} & \textbf{$\Delta$OTS} \\
\midrule

\multirow{4}{*}{Tiny-Aya-Global}
& Off-the-shelf
& 0.103 & --
& 0.824 & --
& 0.421 & -- \\

& Direct MT
& 0.228 & +0.125
& 0.872 & +0.049
& 0.295 & $-$0.126 \\

& Task-Preserved
& 0.232 & +0.129
& 0.874 & +0.050
& 0.446 & +0.025 \\

& \datasettablee{}
& \textbf{0.238} & \textbf{+0.135}
& \textbf{0.877} & \textbf{+0.054}
& \textbf{0.499} & \textbf{+0.078} \\
\midrule

\multirow{4}{*}{Phi-4-Mini-Instruct}
& Off-the-shelf
& 0.091 & --
& 0.827 & --
& 0.258 & -- \\

& Direct MT
& 0.198 & +0.107
& 0.857 & +0.031
& 0.192 & $-$0.066 \\

& Task-Preserved
& 0.200 & +0.109
& 0.859 & +0.033
& 0.262 & +0.004 \\

& \datasettablee{}
& \textbf{0.219} & \textbf{+0.128}
& \textbf{0.867} & \textbf{+0.040}
& \textbf{0.291} & \textbf{+0.033} \\
\midrule

\multirow{4}{*}{Ministral-3-3B-Instruct}
& Off-the-shelf
& 0.118 & --
& 0.829 & --
& 0.253 & -- \\

& Direct MT
& 0.191 & +0.074
& 0.851 & +0.022
& 0.187 & $-$0.066 \\

& Task-Preserved
& 0.193 & +0.076
& 0.855 & +0.026
& 0.277 & +0.025 \\

& \datasettablee{}
& \textbf{0.217} & \textbf{+0.100}
& \textbf{0.867} & \textbf{+0.039}
& \textbf{0.296} & \textbf{+0.044} \\
\midrule

\multirow{4}{*}{Llama-3.2-3B-Instruct}
& Off-the-shelf
& 0.186 & --
& 0.860 & --
& 0.305 & -- \\

& Direct MT
& 0.196 & +0.010
& 0.858 & $-$0.003
& 0.196 & $-$0.109 \\

& Task-Preserved
& 0.201 & +0.015
& 0.860 & +0.000
& 0.263 & $-$0.042 \\

& \datasettablee{}
& \textbf{0.218} & \textbf{+0.032}
& \textbf{0.869} & \textbf{+0.009}
& \textbf{0.310} & \textbf{+0.006} \\
\midrule

\multirow{4}{*}{%
\begin{tabular}{@{}c@{}}
\textbf{Macro Mean} \\
 (across models)
\end{tabular}}
& Off-the-shelf
& 0.124 & --
& 0.835 & --
& 0.309 & -- \\

& Direct MT
& 0.203 & +0.079
& 0.860 & +0.025
& 0.217 & $-$0.092 \\

& Task-Preserved
& 0.207 & +0.083
& 0.862 & +0.027
& 0.312 & +0.003 \\

& \datasettablee{}
& \textbf{0.223} & \textbf{+0.099}
& \textbf{0.870} & \textbf{+0.035}
& \textbf{0.349} & \textbf{+0.040} \\
\bottomrule
\end{tabular}

\caption{
Language-averaged performance for each model, with macro-averages across the four models. Aya ROUGE-L and F\textsc{bert} are averaged over the 36 languages with available reference data, whereas \benchmark{} accuracy is averaged over all 50 evaluated languages. The $\Delta$OTS columns report absolute score differences relative to the corresponding off-the-shelf checkpoint.
}
\label{tab:aggregate-results}
\end{table*}
\section{Results}
\label{sec:results}

\subsection{Reference Similarity versus Instruction Following}
\label{sec:main-results}

As illustrated in Table~\ref{tab:aggregate-results}, Direct MT improves ROUGE-L and F\textsc{bert} on the Aya Evaluation Suite for all evaluated models, except for a small F\textsc{bert} decrease for Llama, while substantially reducing \benchmark{} accuracy for every model. %At the macro-average level, it raises ROUGE-L by 0.079 and F\textsc{bert} by 0.025, but lowers \benchmark{} accuracy by 0.092, showing that higher reference similarity does not necessarily imply better instruction following. 
These results show that greater reference similarity does not necessarily indicate better preservation of the underlying instruction-following task. 

\begin{figure}[t]
    \includegraphics[width=\columnwidth]{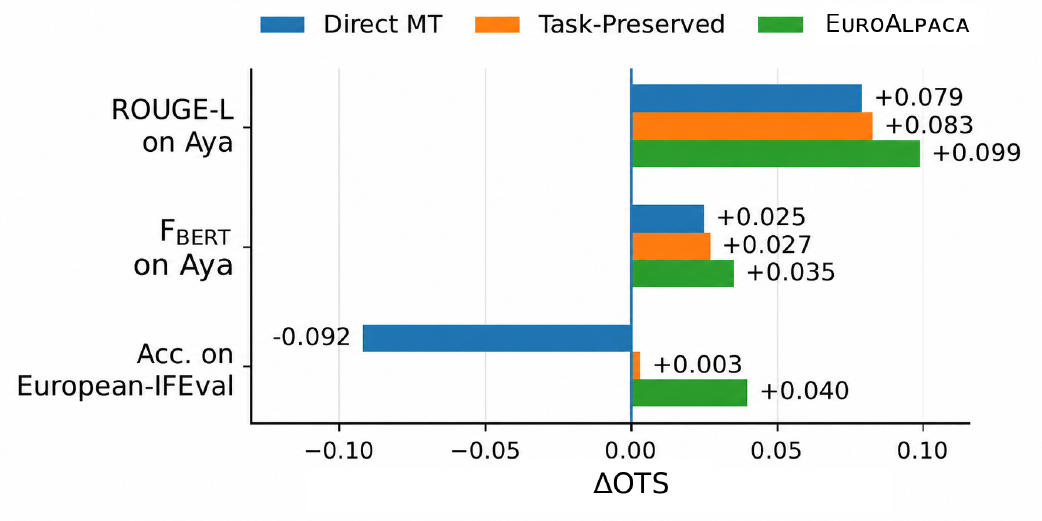}
    \caption{Macro-average $\Delta$OTS values across the four evaluated models}
    %\caption{Performance changes relative to the off-the-shelf checkpoints.} %Direct MT improves reference-based metrics but degrades strict instruction following; task-preserving construction recovers this loss, and final validation provides additional gains.
    \label{fig:metric-changes}
\end{figure}

In contrast, \dataset{} achieves the highest scores across all models and metrics, surpassing Direct MT in the macro-average by 0.020 in Aya ROUGE-L, 0.010 in Aya F\textsc{bert} and 0.132 in \benchmark{} accuracy. As shown by the full language-level results in Appendix~\ref{app:full-results}, \dataset{} outperforms Direct MT in all 200 model--language comparisons. Figure~\ref{fig:metric-changes} summarises changes relative to the off-the-shelf checkpoints.

\subsection{Contributions of Task Preservation and Final Validation}
\label{sec:pipeline-ablation}

The Task-Preserved condition isolates the sample-level gate, field-level preservation and task-level localisation before final validation. As shown in Table~\ref{tab:aggregate-results}, these components improve ROUGE-L, F\textsc{bert} and \benchmark{} accuracy over Direct MT for all models. Notably, Task-Preserved recovers the instruction-following loss caused by Direct MT, slightly exceeds the off-the-shelf mean, and outperforms Direct MT in 199 of the 200 model--language comparisons.

\newpage
Final coherence validation and target-language enforcement yield further gains across all three metrics, increasing Aya ROUGE-L by 0.016, Aya F\textsc{bert} by 0.008 and \benchmark{} accuracy by 0.037 at the macro-average level. \dataset{} outperforms Task-Preserved in 180 of the 200 model--language comparisons, with one tie. These results suggest that task-aware processing prevents most translation-induced task corruption, while final validation addresses complementary errors.

\noindent

\begin{table*}[t]
\centering
\small
\setlength{\tabcolsep}{4pt}
\begin{tabular}{lrrr|rrr|rrr}
\toprule
& & \multicolumn{2}{c}{\textbf{Training data}} & & \multicolumn{2}{c}{\textbf{Training data}} & & \multicolumn{2}{c}{\textbf{Training data}} \\
\cmidrule(lr){3-4} \cmidrule(lr){6-7} \cmidrule(lr){9-10}
\textbf{Eval metric (Macro Avg)} & \textbf{$n$} & \textbf{Okapi} & \textbf{\datasettable{}} & \textbf{$n$} & \textbf{MITS} & \textbf{\datasettable{}} & \textbf{$n$} & \textbf{Bactrian-X} & \textbf{\datasettable{}} \\
\midrule
Aya ROUGE-L $\uparrow$
& 16 & 0.231 & \textbf{0.239}
& 35 & 0.212 & \textbf{0.223}
& 19 & 0.217 & \textbf{0.232} \\
Aya F\textsc{bert} $\uparrow$
& 16 & 0.879 & \textbf{0.879}
& 35 & 0.864 & \textbf{0.870}
& 19 & 0.871 & \textbf{0.878} \\
\benchmark{} Acc. $\uparrow$ 
& 17 & 0.377 & \textbf{0.397}
& 38 & 0.289 & \textbf{0.359}
& 20 & \textbf{0.401} & 0.385 \\
\bottomrule
\end{tabular}
\caption{Pairwise comparison of LoRA fine-tuning runs on Alpaca-derived multilingual instruction resources and on our \dataset{}. For each resource and benchmark, scores are macro-averaged over a fixed shared-language set within each model and then averaged across the four models. $n$ denotes the number of languages shared by the resource and \dataset{} that are also covered by the corresponding evaluation benchmark.}
\label{tab:external-performance}
\end{table*}

\subsection{Comparison with Existing Alpaca-Derived Resources}
\label{sec:external-resources}

Table~\ref{tab:external-performance} reports results on the shared-language subset for each resource and benchmark, averaged across the four models. Because these subsets differ, each comparison is pairwise and should not be used to rank the external resources against one another. \dataset{} outperforms Okapi and MITS on all reported metrics, except for a tie with Okapi on Aya F\textsc{bert}. On \benchmark{}, it improves accuracy by 0.020 over Okapi and by 0.070 over MITS. Compared with Bactrian-X, \dataset{} achieves higher Aya scores but lower \benchmark{} accuracy on the 20-language intersection (0.385 vs.\ 0.401).

\newpage
We also examine cross-language sample alignment in Table~\ref{tab:external-alignment}. % shows that \dataset{}, Okapi and Bactrian-X permit sample-level matching across languages. 
Okapi and \dataset{} obtain similar embedding-similarity scores (0.867 vs. 0.860), whereas Bactrian-X is less closely aligned (0.800). MITS does not expose identifiers that permit this analysis. We interpret embedding similarity as a measure of cross-language consistency rather than task validity or overall dataset quality.

\begin{table}[t]
\centering
\small
\setlength{\tabcolsep}{5pt}
\begin{tabular}{lcc}
\toprule
\textbf{Resource} &
\shortstack{\textbf{Sample-level}\\\textbf{alignment}} &
\shortstack{\textbf{Embedding}\\\textbf{similarity}} \\
\midrule
Okapi      & Yes & 0.867 \\
\dataset{} & Yes & 0.860 \\
Bactrian-X & Yes & 0.800 \\
MITS       & No  & --   \\
\bottomrule
\end{tabular}
\caption{
Sample-level alignment indicates whether corresponding records can be matched across languages. Embedding similarity is the mean cosine similarity between BGE-M3 representations of aligned instruction--input--output records over the common 11-language intersection of \dataset{}, Okapi and Bactrian-X.
}
\label{tab:external-alignment}
\end{table}
\section{Discussion}

The central finding is the divergence between reference- and constraint-based evaluation. Direct MT improves ROUGE-L and F\textsc{bert} on Aya but causes an absolute decrease of 0.092 in \benchmark{} accuracy. Thus, models trained on directly translated supervision can produce outputs that are closer to the references while becoming less reliable at following verifiable instructions. %, highly aligned with findings from~\citealp{DBLP:journals/corr/abs-2603-10351}.
Multilingual instruction data should consequently be evaluated through both target-language response quality and verifiable instruction following.

The ablation results clarify how the pipeline addresses this problem. The Task-Preserved condition yields an accuracy gain of 0.095 over Direct MT, showing that sample-level routing, field preservation and task-level localisation prevent much of the translation-induced task corruption. Final validation adds a further gain of 0.037 and also improves Aya scores, indicating that cross-field consistency and target-language enforcement contribute beyond the initial generation policy.
The two stages are therefore complementary.

Comparisons with external resources provide a broader view of the results. \dataset{} outperforms Okapi and MITS on their respective shared-language subsets, whereas Bactrian-X achieves higher \benchmark{} accuracy on its language set, possibly due to its direct target-language response generation approach. At the same time, \dataset{} achieves cross-language embedding similarity close to Okapi's and higher than Bactrian-X's. These results show that near-parallel alignment is compatible with competitive downstream performance. However, differences in construction procedures prevent attributing the observed performance gaps solely to task-preserving localisation.
\section{Conclusion}
We introduce \dataset{}, an LLM-guided pipeline and near-parallel resource for task-preserving instruction-data localisation across 50 European languages and regional varieties. The pipeline combines sample- and field-level translation policy predictions, policy-guided MT, task-level rewriting and final target-language and cross-field coherence validation. Across experiments with four LLMs, direct MT improves reference similarity on the Aya Evaluation Suite but reduces \benchmark{} accuracy by 29.8\% relative to the unadapted baseline. In contrast, \dataset{} improves accuracy by 12.9\% over the same baseline, fully recovering the loss caused by direct MT, while also improving open-ended generation metrics and outperforming direct MT in all 200 model--language pairs on instruction following. These findings underscore the importance of task-preserving localisation and highlight the need for complementary reference- and constraint-based evaluation.

\section*{Limitations}
This study covers only European languages and regional varieties, although the proposed approach could be applied to languages from other regions. The semantic labels, predicted policies, localised examples and validation edits are model-assisted and may therefore inherit model errors or biases. A key limitation of our pipeline is its reliance on Gemma-4 (\texttt{google/gemma-4-31B-it}) for translation policy prediction, rewriting and validation. Consequently, the quality of task-preserving localisations in low-resource varieties (such as Latgalian, Lombard or Friulian) is bounded by the LLM's pre-existing exposure to and knowledge of these languages.
Although we manually inspected a small sample of outputs, a larger and more systematic error analysis across languages and processing stages would provide a stronger assessment of annotation, localisation and validation quality. We report results from a single training seed for each model-language–condition combination and therefore do not estimate optimisation variance. Comparisons with external datasets are restricted to matched model--language cells and should not be interpreted as comprehensive evaluations of those resources.

\section*{Data Availability Statement}
At the time of submission, \dataset{} and \benchmark{} are not publicly available in order to preserve anonymity during the double-blind review process. Upon acceptance, we plan to release both datasets, the complete data-construction code and the intermediate annotations produced at each stage of the pipeline. These materials will include task and domain labels, sample- and field-level translation policies, examples produced through task-level localisation, validation decisions and post-edit metadata. We also plan to provide the prompt templates, processing configurations, provenance manifests, evaluation scripts and documentation required to reproduce the datasets and experiments.

\section*{AI-Generated Content Disclosure}
We used a generative AI to assist in paraphrasing, improving clarity and grammar in parts of the manuscript and coding assistance. All generated content was reviewed and validated by the authors.

\section*{Acknowledgment}
This work has received funding from the European Union's Horizon Europe research and innovation programme under the project ELOQUENCE (Grant Agreement No. 101135916). This work was supported by computational resources from the EuroHPC Joint Undertaking under the EuroHPC AI Factory grant EHPC-AIF-2026LS01-004.
Carlos Escolano is funded by Grant  LLM4SL (PID2024-157855OA-C33) MICIU/AEI/10.13039/501100011033 and by “ERDF/EU”.

\bibliography{references}

\newpage
\appendix
%\section{Dataset and Policy Statistics}
%\label{app:statistics}

% Reformatted appendix fragment.
% Assumes the main document already loads graphicx and booktabs and defines \dataset{}.
% All tables use the document's standard numeric table counter.

\clearpage
\onecolumn

% Fully justified manual captions for fixed, non-floating material.
% Resetting the paragraph skips here prevents an enclosing \centering command
% from centering the caption text.
\newcommand{\appcaptionparagraph}[1]{%
  \par\smallskip
  \begingroup
  \footnotesize
  \leftskip=0pt\relax
  \rightskip=0pt\relax
  \parfillskip=0pt plus 1fil\relax
  \noindent #1\par
  \endgroup
}
\newcommand{\appmanualfigcaption}[2]{%
  \refstepcounter{figure}\label{#2}%
  \appcaptionparagraph{Figure~\thefigure: #1}%
}
\newcommand{\appmanualtabcaption}[2]{%
  \refstepcounter{table}\label{#2}%
  \appcaptionparagraph{Table~\thetable: #1}%
}

\section{Dataset and Policy Statistics}
\label{app:statistics}

\subsection{Semantic Annotation of \datasettitle{}}
\label{distributions}
\vspace{0.5\baselineskip}
% -----------------------------------------------------------------------------
% Page 1: fixed two-column layout.
% Left: task taxonomy, domain taxonomy, Figure 4.
% Right: Figure 5, localisation-reason table, field-policy table.
% -----------------------------------------------------------------------------
\noindent
\begin{minipage}[t]{0.485\textwidth}
\vspace{0pt}

% Task taxonomy table.
\begin{center}
\scriptsize
\setlength{\tabcolsep}{3pt}
\renewcommand{\arraystretch}{0.95}
\begin{tabular}{@{}p{0.48\linewidth}p{0.48\linewidth}@{}}
\toprule
\textbf{TASK TAXONOMY} \\
\midrule
Creative writing          & Question answering \\
Reasoning                 & Information extraction \\
Classification            & Paraphrasing \\
Procedural instructions   & Coding \\
Summarisation             & Style transfer \\
Evaluation                & Translation \\
Other                     & \\
\bottomrule
\end{tabular}
\end{center}
\appmanualtabcaption{Task-label taxonomy used for semantic annotation of the source examples.}{tab:task-taxonomy}

\vspace{3\baselineskip}

% Domain taxonomy table.
\begin{center}
\scriptsize
\setlength{\tabcolsep}{3pt}
\renewcommand{\arraystretch}{0.95}
\begin{tabular}{@{}p{0.48\linewidth}p{0.48\linewidth}@{}}
\toprule
\textbf{DOMAIN TAXONOMY} \\
\midrule
General        & Creative \\
Technology     & Business \\
Science        & Humanities \\
Mathematics    & Education \\
Programming    & Social science \\
Medicine       & Law \\
\bottomrule
\end{tabular}
\end{center}
\appmanualtabcaption{Domain-label taxonomy used for semantic annotation of the source examples.}{tab:domain-taxonomy}

\vspace{3\baselineskip}

% Figure 4 below both taxonomy tables.
\begin{center}
\includegraphics[width=\linewidth]{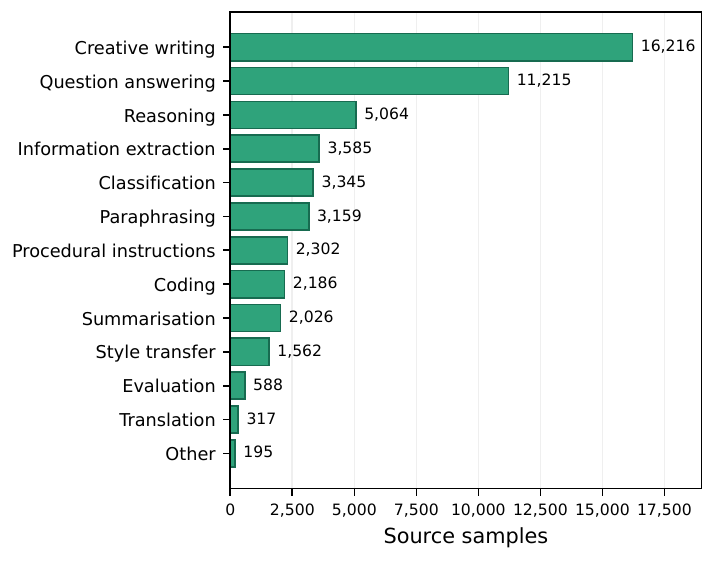}
\end{center}
\appmanualfigcaption{Primary task-label distribution across the 51,760 English source examples, as assigned by the LLM using the predefined task taxonomy.}{fig:task-distribution}

\end{minipage}
\hfill
\begin{minipage}[t]{0.485\textwidth}
\vspace{0pt}

% Figure 5 starts the right column.
\begin{center}
\includegraphics[width=0.94\linewidth]{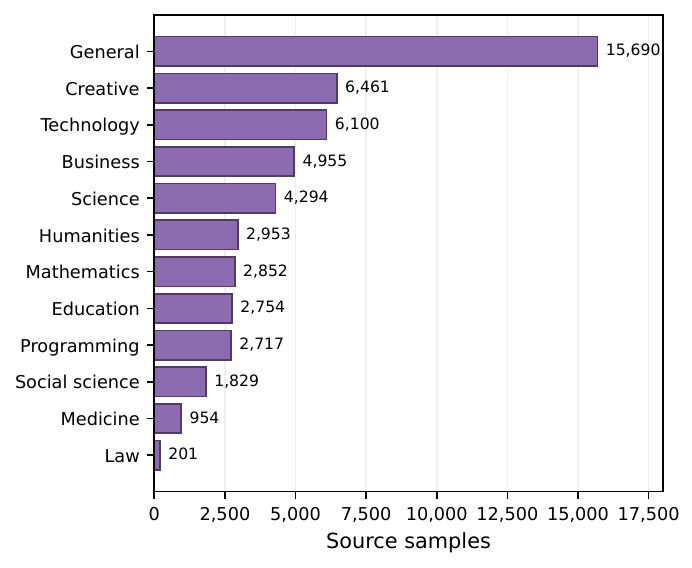}
\end{center}
\appmanualfigcaption{Primary domain-label distribution across the 51,760 English source examples, as assigned by the LLM using the predefined domain taxonomy.}{fig:domain-distribution}

\vspace{3\baselineskip}

% Table 9.

\begin{center}
\scriptsize
\setlength{\tabcolsep}{3pt}
\renewcommand{\arraystretch}{0.96}
\begin{tabular}{@{}p{0.64\linewidth}rr@{}}
\toprule
\textbf{Task-level localisation reason} &
\textbf{Count} &
\textbf{\%} \\
\midrule
English grammar or spelling        & 4,694 & 42.3 \\
Letter or character constraint     & 2,369 & 21.4 \\
Language-specific task             & 1,165 & 10.5 \\
Word/line-count constraint         & 1,053 & 9.5 \\
Rhyme or sound constraint          & 644   & 5.8 \\
Idiom or proverb                   & 411   & 3.7 \\
Wordplay/pun/anagram/palindrome    & 387   & 3.5 \\
Other non-literal case             & 220   & 2.0 \\
Mnemonic or initial letters        & 85    & 0.8 \\
Source output violates instruction & 60    & 0.5 \\
\midrule
Total                     & 11,088 & 100.0 \\
\bottomrule
\end{tabular}
\end{center}
\appmanualtabcaption{Reasons for routing 11,088 examples to task-level localisation.}{tab:localisation-reasons}

\vspace{3\baselineskip}

% Table 10.
\begin{center}
\scriptsize
\setlength{\tabcolsep}{5pt}
\renewcommand{\arraystretch}{1.0}
\begin{tabular}{@{}lrr@{}}
\toprule
\textbf{Field} & \textbf{Translate} & \textbf{Preserve} \\
\midrule
Instruction & 40,174 & 498 \\
Input       & 7,328  & 33,344 \\
Output      & 30,179 & 10,493 \\
\bottomrule
\end{tabular}
\end{center}
\appmanualtabcaption{Field policies for the 40,672 examples that pass the whole-sample translation gate. Empty inputs are counted as preserved.}{tab:field-policies}

\end{minipage}

% -----------------------------------------------------------------------------
% Page 2: subsection title and Table 11 on the same page.
% -----------------------------------------------------------------------------
\clearpage
\subsection{Examples Across Task-Preserving Handling Regimes}
\label{app:broken-direct-mt}

\vspace{3\baselineskip}

\begin{center}
\scriptsize
\begin{tabular}{p{0.05\textwidth}p{0.14\textwidth}p{0.24\textwidth}p{0.24\textwidth}p{0.24\textwidth}}
\toprule
\textbf{ID} & \textbf{Regime} & \textbf{Source example} & \textbf{Direct field-wise translation} & \textbf{Task-preserving target example} \\
\midrule
139
& Task-equivalent rewriting in the target-language: Source- language form
& \textbf{English} \newline
\textbf{Instruction:} Edit the following sentence to correct the grammar and syntax errors. \newline
\textbf{Input:} You was late to the party. \newline
\textbf{Output:} You were late to the party.
& \textbf{French} \newline
\textbf{Instruction:} Modifie la phrase suivante pour corriger les erreurs de grammaire et de syntaxe. \newline
\textbf{Input:} Tu étais en retard à la fête. \newline
\textbf{Output:} Tu étais en retard à la fête.
& \textbf{French} \newline
\textbf{Instruction:} Modifie la phrase suivante pour corriger les erreurs de grammaire et de syntaxe. \newline
\textbf{Input:} Tu était en retard à la fête. \newline
\textbf{Output:} Tu étais en retard à la fête. \\
\addlinespace[0.5em]
\multicolumn{2}{r}{\emph{Failure and handling:}}
& \multicolumn{3}{p{0.74\textwidth}}{Direct translation can turn the erroneous source sentence into a grammatical target-language sentence, making the input and output identical. The localised sample must introduce a comparable target-language error and provide its correction.} \\
\midrule
1480
& Task-equivalent rewriting in the target-language: Rhyme / phonology
& \textbf{English} \newline
\textbf{Instruction:} Generate a list of words that rhyme with ``cat''. \newline
\textbf{Output:} Bat, rat, hat, mat, pat, ...
& \textbf{German} \newline
\textbf{Instruction:} Erstelle eine Liste von Wörtern, die sich auf ``Katze'' reimen. \newline
\textbf{Output:} Fledermaus, Ratte, Hut, Matte, Klaps, ...
& \textbf{German} \newline
\textbf{Instruction:} Erstelle eine Liste von Wörtern, die sich auf ``Haus'' reimen. \newline
\textbf{Output:} Maus, Klaus, Applaus, hinaus. \\
\addlinespace[0.5em]
\multicolumn{2}{r}{\emph{Failure and handling:}}
& \multicolumn{3}{p{0.74\textwidth}}{Rhyme is tied to target-language phonology. The cue word and answer set must be selected in the target language rather than translated word by word.} \\
\midrule
218
& Task-equivalent rewriting in the target-language: Structural constraint
& \textbf{English} \newline \textbf{Instruction:} Generate a haiku poem. \newline
\textbf{Output:} Fallen crimson leaves / Sighing softly in the breeze / Nature's tranquil peace.
& \textbf{Catalan} \newline
\textbf{Instruction:} Genera un poema haiku. \newline
\textbf{Output:} Fulles carmesines caigudes / sospirant suaument a la brisa / pau tranquil.la de la natura.
& \textbf{Catalan} \newline
\textbf{Instruction:} Genera un haiku. \newline
\textbf{Output:} Fulles al vent / la tarda calla lenta / pau de tardor. \\
\addlinespace[0.5em]
\multicolumn{2}{r}{\emph{Failure and handling:}}
& \multicolumn{3}{p{0.74\textwidth}}{Line, syllable and poetic-form constraints are not preserved by literal translation. The poem must be regenerated so that the target text satisfies the intended form.} \\
\midrule
115
& Task-equivalent rewriting in the target-language: Idiom
& \textbf{English} \newline \textbf{Instruction:} What is the meaning of the following idiom? \newline
\textbf{Input:} It's raining cats and dogs. \newline
\textbf{Output:} It means that it is raining very heavily.
& \textbf{Danish} \newline
\textbf{Instruction:} Hvad betyder følgende idiom? \newline
\textbf{Input:} Det regner katte og hunde. \newline
\textbf{Output:} Det betyder, at det regner meget kraftigt.
& \textbf{Danish} \newline
\textbf{Instruction:} Hvad betyder følgende udtryk? \newline
\textbf{Input:} Det står ned i stænger. \newline
\textbf{Output:} Det betyder, at det regner meget kraftigt. \\
\addlinespace[0.5em]
\multicolumn{2}{r}{\emph{Failure and handling:}}
& \multicolumn{3}{p{0.74\textwidth}}{The literal translation of the English idiom is not a natural Danish expression. The source idiom should be replaced by a target-language idiom with the same function.} \\
\midrule
744
& Task-equivalent rewriting in the target-language: Mnemonic / initials
& \textbf{English} \newline
\textbf{Instruction:} Create a mnemonic to remember the eight planets. \newline
\textbf{Output:} ``My Very Eager Mother Just Served Us Nachos''...
& \textbf{Italian} \newline
\textbf{Instruction:} Crea una frase mnemonica per ricordare gli otto pianeti. \newline
\textbf{Output:} ``Mia Madre Molto Entusiasta Ha Appena Servito Nachos''...
& \textbf{Italian} \newline
\textbf{Instruction:} Crea una frase mnemonica per ricordare gli otto pianeti. \newline
\textbf{Output:} ``Mia Vecchia Terra Mi Guarda Sempre Un Nuvolone'', per Mercurio, Venere, Terra, Marte, Giove, Saturno, Urano e Nettuno. \\
\addlinespace[0.5em]
\multicolumn{2}{r}{\emph{Failure and handling:}}
& \multicolumn{3}{p{0.74\textwidth}}{The English mnemonic encodes initial-letter constraints. The target sample must create a new mnemonic compatible with the target-language planet names.} \\
\midrule
2575
& Task-equivalent rewriting in the target-language: Letter-count constraint
& \textbf{English} \newline  \textbf{Instruction:} Find a five-letter word that begins with `d' and ends with `s'. \newline
\textbf{Output:} Dress
& \textbf{Spanish} \newline
\textbf{Instruction:} Encuentra una palabra de cinco letras que empiece por `d' y termine en `s'. \newline
\textbf{Output:} Vestido
& \textbf{Spanish} \newline
\textbf{Instruction:} Encuentra una palabra de cinco letras que empiece por `d' y termine en `s'. \newline
\textbf{Output:} Dosis \\
\addlinespace[0.5em]
\multicolumn{2}{r}{\emph{Failure and handling:}}
& \multicolumn{3}{p{0.74\textwidth}}{The English answer satisfies the letter constraint, but its direct translation does not. The localised sample must choose a target-language answer that satisfies the same formal constraint.} \\
\bottomrule
\end{tabular}
\end{center}
\appmanualtabcaption{Examples requiring task-level localisation. The fourth column illustrates the failure cases, while the final column shows a task-preserving target-language construction.}{tab:appendix-localisation-examples}

% -----------------------------------------------------------------------------
% Page 3: Table 12 only.
% -----------------------------------------------------------------------------
\clearpage

\vspace{5\baselineskip}
\begin{center}
\scriptsize
\begin{tabular}{p{0.05\textwidth}p{0.14\textwidth}p{0.24\textwidth}p{0.24\textwidth}p{0.24\textwidth}}
\toprule
\textbf{ID} & \textbf{Regime} & \textbf{Source example} & \textbf{Direct field-wise translation} & \textbf{Task-preserving target example} \\
\midrule
66
& Direct translation
& \textbf{English} \newline
\textbf{Instruction:} Find the main idea of the following passage. \newline
\textbf{Input:} Climate change is an increasingly contentious global issue... \newline
\textbf{Output:} The main idea is that climate change is severely impacting the environment...
& \textbf{Finnish} \newline
\textbf{Instruction:} Etsi seuraavan tekstin pääajatus. \newline
\textbf{Input:} Ilmastonmuutos on yhä kiistanalaisempi maailmanlaajuinen kysymys... \newline
\textbf{Output:} Tekstin pääajatus on, että ilmastonmuutos vaikuttaa vakavasti ympäristöön...
& \textbf{Finnish} \newline
\textbf{Instruction:} Etsi seuraavan tekstin pääajatus. \newline
\textbf{Input:} Ilmastonmuutos on yhä kiistanalaisempi maailmanlaajuinen kysymys... \newline
\textbf{Output:} Tekstin pääajatus on, että ilmastonmuutos vaikuttaa vakavasti ympäristöön... \\
\addlinespace[0.5em]
\multicolumn{2}{r}{\emph{Failure and handling:}}
& \multicolumn{3}{p{0.74\textwidth}}{This is a direct-translation case. All non-empty fields can be translated without changing the task relation. The direct and task-preserving target examples are therefore identical.} \\
\midrule
192
& Direct translation
& \textbf{English} \newline
\textbf{Instruction:} Compose an informative sentence explaining the given term. \newline
\textbf{Input:} Green bonds \newline
\textbf{Output:} Green bonds are debt instruments issued to raise funds for environmentally friendly projects.
& \textbf{Romanian} \newline
\textbf{Instruction:} Scrie o propoziție informativă care explică termenul dat. \newline
\textbf{Input:} Obligațiuni verzi \newline
\textbf{Output:} Obligațiunile verzi sunt instrumente de datorie emise pentru a strânge fonduri pentru proiecte ecologice.
& \textbf{Romanian} \newline
\textbf{Instruction:} Scrie o propoziție informativă care explică termenul dat. \newline
\textbf{Input:} Obligațiuni verzi \newline
\textbf{Output:} Obligațiunile verzi sunt instrumente de datorie emise pentru a strânge fonduri pentru proiecte ecologice. \\
\addlinespace[0.5em]
\multicolumn{2}{r}{\emph{Failure and handling:}}
& \multicolumn{3}{p{0.74\textwidth}}{This is a direct-translation case. The instruction, input, and output express ordinary semantic content, so translating all non-empty fields preserves the task relation.} \\
\midrule
98
& Field-preserving
& \textbf{English} \newline
\textbf{Instruction:} Classify the following statement as true or false. \newline
\textbf{Input:} The Supreme Court is the highest court in the US. \newline
\textbf{Output:} True
& \textbf{Spanish} \newline
\textbf{Instruction:} Clasifica la siguiente afirmación como verdadera o falsa. \newline
\textbf{Input:} El Tribunal Supremo es el tribunal más alto de EE. UU. \newline
\textbf{Output:} Es cierto.
& \textbf{Spanish} \newline
\textbf{Instruction:} Classify the following statement as true or false. \newline
\textbf{Input:} El Tribunal Supremo es el tribunal más alto de EE. UU. \newline
\textbf{Output:} True \\
\addlinespace[0.5em]
\multicolumn{2}{r}{\emph{Failure and handling:}}
& \multicolumn{3}{p{0.74\textwidth}}{The task defines a fixed label set. Translating the labels into natural Spanish changes the required answer space, so the instruction label set and the output label must be preserved verbatim.} \\
\midrule
2092
& Field-preserving
& \textbf{English} \newline
\textbf{Instruction:} Construct a valid SQL statement. \newline
\textbf{Input:} Retrieve the names and ages of all the students who live in Brisbane. \newline
\textbf{Output:} \texttt{SELECT name, age FROM students WHERE city = `Brisbane';}
& \textbf{Albanian} \newline
\textbf{Instruction:} Ndërto një deklaratë të vlefshme SQL. \newline
\textbf{Input:} Merr emrat dhe moshat e të gjithë studentëve që jetojnë në Brisbane. \newline
\textbf{Output:} \texttt{ZGJIDH emri, mosha NGA studentet KU qyteti = `Brisbane';}
& \textbf{Albanian} \newline
\textbf{Instruction:} Ndërto një deklaratë të vlefshme SQL. \newline
\textbf{Input:} Retrieve the names and ages of all the students who live in Brisbane. \newline
\textbf{Output:} \texttt{SELECT name, age FROM students WHERE city = `Brisbane';} \\
\addlinespace[0.5em]
\multicolumn{2}{r}{\emph{Failure and handling:}}
& \multicolumn{3}{p{0.74\textwidth}}{The output is executable SQL. Translating SQL keywords, column names, table names or string literals can make the query invalid or change its semantics, so the SQL statement is preserved.} \\
\midrule
38024
& Field-preserving
& \textbf{English} \newline
\textbf{Instruction:} Translate English text to French. \newline
\textbf{Input:} The sun was shining brightly in the sky. \newline
\textbf{Output:} Le soleil brillait fort dans le ciel.
& \textbf{Welsh} \newline
\textbf{Instruction:} Cyfieithwch destun Saesneg i'r Ffrangeg. \newline
\textbf{Input:} Roedd yr haul yn disgleirio'n llachar yn yr awyr. \newline
\textbf{Output:} Roedd yr haul yn disgleirio'n gryf yn yr awyr.
& \textbf{Welsh} \newline
\textbf{Instruction:} Cyfieithwch destun Saesneg i'r Ffrangeg. \newline
\textbf{Input:} The sun was shining brightly in the sky. \newline
\textbf{Output:} Le soleil brillait fort dans le ciel. \\
\addlinespace[0.5em]
\multicolumn{2}{r}{\emph{Failure and handling:}}
& \multicolumn{3}{p{0.74\textwidth}}{The source and target languages are part of the task definition. Translating every field changes the English--French translation pair into Welsh text, so the source sentence and fixed-language output must be preserved.} \\
\midrule
3798
& Field-preserving
& \textbf{English} \newline
\textbf{Instruction:} Complete the following sentence by filling in the \texttt{<mask>}. \newline
\textbf{Input:} I wanted to throw a party but the \texttt{<mask>} became an obstacle. \newline
\textbf{Output:} I wanted to throw a party but the pandemic became an obstacle.
& \textbf{Icelandic} \newline
\textbf{Instruction:} Ljúktu við eftirfarandi setningu með því að fylla inn í \texttt{<gríma>}. \newline
\textbf{Input:} Ég vildi halda veislu en \texttt{<gríma>} varð hindrun. \newline
\textbf{Output:} Ég vildi halda veislu en faraldurinn varð hindrun.
& \textbf{Icelandic} \newline
\textbf{Instruction:} Ljúktu við eftirfarandi setningu með því að fylla inn í \texttt{<mask>}. \newline
\textbf{Input:} I wanted to throw a party but the \texttt{<mask>} became an obstacle. \newline
\textbf{Output:} I wanted to throw a party but the pandemic became an obstacle. \\
\addlinespace[0.5em]
\multicolumn{2}{r}{\emph{Failure and handling:}}
& \multicolumn{3}{p{0.74\textwidth}}{The mask token is part of the task format. Translating it changes the literal token that must be filled, so the token and masked sentence are preserved.} \\
\midrule
32293
& Field-preserving
& \textbf{English} \newline
\textbf{Instruction:} Translate the following sentence into Latin. \newline
\textbf{Input:} He is happy. \newline
\textbf{Output:} Is laetus est.
& \textbf{Czech} \newline
\textbf{Instruction:} Přelož následující větu do latiny. \newline
\textbf{Input:} Je šťastný. \newline
\textbf{Output:} Je šťastný.
& \textbf{Czech} \newline
\textbf{Instruction:} Přelož následující větu do latiny. \newline
\textbf{Input:} Je šťastný. \newline
\textbf{Output:} Is laetus est. \\
\addlinespace[0.5em]
\multicolumn{2}{r}{\emph{Failure and handling:}}
& \multicolumn{3}{p{0.74\textwidth}}{The instruction and source sentence can be localised, but the output is the requested Latin translation. Translating the output into the target language removes the answer, so the output field is preserved.} \\
\bottomrule
\end{tabular}
\end{center}
\appmanualtabcaption{Examples of direct and field-preserving translation. Direct field-wise translation preserves the task in the first two examples, but later examples require preserving task-critical fields such as labels, code, literal tokens and fixed-language outputs.}{tab:appendix-preservation-examples}

\clearpage
\twocolumn
\clearpage
\clearpage
\onecolumn
\section{Language Coverage and Analysis Tiers}
\label{app:languages}

\begin{multicols}{2}
\dataset{} comprises an annotated English source collection and
near-parallel versions in 50 European languages and regional varieties.
Table~\ref{tab:language-coverage} lists the target languages alphabetically,
together with the ISO tags used in the dataset and experiments and their
assigned resource tiers.

The high-, medium- and low-resource categories are internal analysis
strata reflecting the relative availability of digital resources considered
during dataset construction. They should not be interpreted as measures of
language vitality, speaker population or sociolinguistic status.
\end{multicols}

\centering
\footnotesize
\setlength{\tabcolsep}{4pt}
\renewcommand{\arraystretch}{0.92}

\begin{tabular}{@{}lll@{}}
\toprule
\textbf{ISO tag} & \textbf{Language} & \textbf{Resource tier} \\
\midrule
\texttt{sq}  & Albanian            & Medium \\
\texttt{ast} & Asturian            & Low \\
\texttt{eu}  & Basque              & Medium \\
\texttt{bs}  & Bosnian             & Medium \\
\texttt{bg}  & Bulgarian           & High \\
\texttt{ca}  & Catalan             & High \\
\texttt{hr}  & Croatian            & Medium \\
\texttt{cs}  & Czech               & High \\
\texttt{da}  & Danish              & High \\
\texttt{nl}  & Dutch               & High \\
\texttt{et}  & Estonian            & Medium \\
\texttt{fo}  & Faroese             & Low \\
\texttt{fi}  & Finnish             & High \\
\texttt{fr}  & French              & High \\
\texttt{fur} & Friulian            & Low \\
\texttt{gl}  & Galician            & Medium \\
\texttt{ka}  & Georgian            & Medium \\
\texttt{de}  & German              & High \\
\texttt{el}  & Greek               & High \\
\texttt{hu}  & Hungarian           & High \\
\texttt{is}  & Icelandic           & Medium \\
\texttt{ga}  & Irish               & Medium \\
\texttt{it}  & Italian             & High \\
\texttt{ltg} & Latgalian           & Low \\
\texttt{lv}  & Latvian             & Medium \\
\texttt{lij} & Ligurian            & Low \\
\texttt{lt}  & Lithuanian          & High \\
\texttt{lmo} & Lombard             & Low \\
\texttt{lb}  & Luxembourgish       & Low \\
\texttt{mk}  & Macedonian          & Medium \\
\texttt{mt}  & Maltese             & Medium \\
\texttt{nob} & Norwegian Bokmål    & High \\
\texttt{nno} & Norwegian Nynorsk   & Low \\
\texttt{oc}  & Occitan             & Low \\
\texttt{pl}  & Polish              & High \\
\texttt{pt}  & Portuguese          & High \\
\texttt{ro}  & Romanian            & High \\
\texttt{ru}  & Russian             & High \\
\texttt{sc}  & Sardinian           & Low \\
\texttt{gd}  & Scottish Gaelic     & Low \\
\texttt{sr}  & Serbian             & Medium \\
\texttt{scn} & Sicilian            & Low \\
\texttt{szl} & Silesian            & Low \\
\texttt{sk}  & Slovak              & High \\
\texttt{sl}  & Slovenian           & Medium \\
\texttt{es}  & Spanish             & High \\
\texttt{sv}  & Swedish             & High \\
\texttt{uk}  & Ukrainian           & High \\
\texttt{vec} & Venetian            & Low \\
\texttt{cy}  & Welsh               & Medium \\
\bottomrule
\end{tabular}
\captionof{table}{
Language coverage and MADLAD-400-based resource tiers for \dataset{}.
Tiers are defined using clean monolingual token counts: high
($\geq 5$ billion), medium ($100$ million to $<5$ billion) and low
($<100$ million or no separate clean partition). The classification
contains 21 high-, 15 medium- and 14 low-resource languages and is used
only for aggregate analysis.
}
\label{tab:language-coverage}

\clearpage

\subsection{\benchmarktitle{} Provenance}
\label{app:ifeval-provenance}
\vspace{2\baselineskip}

\begin{center}
\begin{minipage}{0.96\textwidth}
\centering
\small
\setlength{\tabcolsep}{6pt}
\renewcommand{\arraystretch}{1.05}

% The source column is wider than in the original table.
\begin{tabular}{
  p{0.18\textwidth}
  p{0.40\textwidth}
  p{0.33\textwidth}
}
\toprule
\textbf{Tier} & \textbf{Languages} & \textbf{Hugging Face Dataset ID} \\
\midrule

\multirow{10}{=}{
  \makecell[l]{
    Native-speaker or\\
    professional translator\\
    (human-curated)
  }
}
& ca
& \href{https://huggingface.co/datasets/projecte-aina/IFEval_ca}
        {projecte-aina/IFEval\_ca%\footnote{\url{ https://huggingface.co/datasets/projecte-aina/IFEval_ca}}
        } \\
\cmidrule(lr){2-3}

& da
& \href{https://huggingface.co/datasets/danish-foundation-models/ifeval-da}
        {danish-foundation-models/ifeval-da} \\
\cmidrule(lr){2-3}

& de
& \href{https://huggingface.co/datasets/jzhang86/de_ifeval}
        {jzhang86/de\_ifeval} \\
\cmidrule(lr){2-3}

& el
& \href{https://huggingface.co/datasets/ilsp/ifeval_greek}
        {ilsp/ifeval\_greek} \\
\cmidrule(lr){2-3}

& es
& \href{https://huggingface.co/datasets/BSC-LT/IFEval_es}
        {BSC-LT/IFEval\_es} \\
\cmidrule(lr){2-3}

& et
& \href{https://huggingface.co/datasets/tartuNLP/ifeval_et}
        {tartuNLP/ifeval\_et} \\
\cmidrule(lr){2-3}

& eu
& \href{https://huggingface.co/datasets/HiTZ/ifeval_eu}
        {HiTZ/ifeval\_eu} \\
\cmidrule(lr){2-3}

& fi
& \href{https://huggingface.co/datasets/LumiOpen/ifeval_mt/tree/main/fi}
        {LumiOpen/ifeval\_mt (Finnish)} \\
\cmidrule(lr){2-3}

& gl
& \href{https://huggingface.co/datasets/HiTZ/ifeval_gl}
        {HiTZ/ifeval\_gl} \\
\cmidrule(lr){2-3}

& sv
& \href{https://huggingface.co/datasets/LumiOpen/ifeval_mt/tree/main/sv}
        {LumiOpen/ifeval\_mt (Swedish)} \\

\midrule

Localised and verified
& cs, fr, hu, it, nl, pl, pt, ro, ru, uk
& \href{https://huggingface.co/datasets/AIDC-AI/Marco-Bench-MIF}
        {AIDC-AI/Marco-Bench-MIF} \\

\midrule

Machine-translated with localisation
& ast, bg, bs, cy, fo, fur, ga, gd, hr, is, ka, lb, lij, lmo,
  lt, ltg, lv, mk, mt, nno, nob, oc, sc, scn, sk, sl, sq, sr, szl, vec
& \href{https://huggingface.co/datasets/danish-foundation-models/multi-ifeval}
        {danish-foundation-models/multi-ifeval} \\

\bottomrule
\end{tabular}

\captionsetup{
  justification=justified,
  singlelinecheck=false
}
\captionof{table}{
Language-level provenance summary for \benchmark{} data.
The released manifest stores the exact URL, source tier, missing-sample
count and source-language answer-constraint count for each language.}
\label{tab:ifeval-provenance-summary}

\end{minipage}
\end{center}

\clearpage
\twocolumn
\clearpage
\clearpage
\onecolumn
\section{Complete Language-Level Results}
\label{app:full-results}
%\vspace{-5.5em}
%\begin{multicols}{2}
%This appendix reports the complete per-language results for all four models and supervision conditions. \benchmark{} is evaluated across all 50 target languages, whereas the Aya Evaluation Suite covers 36 non-English target languages. English scores are retained where available but are excluded from Aya macro-averages. A dash denotes an unavailable result.  External-dataset comparisons are reported separately in Appendix~\ref{app:additional-results}.
%\end{multicols}

\vspace{1.35\baselineskip}
\subsection{\benchmarktitle{}: Accuracy after Language-Specific Fine-Tuning}
\vspace{0.85\baselineskip}
\centering
\tiny
\setlength{\tabcolsep}{1.35pt}
\renewcommand{\arraystretch}{0.90}
\resizebox{0.98\textwidth}{!}{%
\begin{tabular}{llrrrrrrrrrrrrrrrrrrrr}
\toprule
\textbf{Model} & \textbf{Condition} & \textbf{ast} & \textbf{bg} & \textbf{bs} & \textbf{ca} & \textbf{cs} & \textbf{cy} & \textbf{da} & \textbf{de} & \textbf{el} & \textbf{es} & \textbf{et} & \textbf{eu} & \textbf{fi} & \textbf{fo} & \textbf{fr} & \textbf{fur} & \textbf{ga} & \textbf{gd} & \textbf{gl} & \textbf{hr} \\
\midrule
Tiny-Aya-Global & Off-the-shelf & .392 & .531 & .442 & .506 & .460 & .454 & .549 & .468 & .497 & .508 & .468 & .410 & .475 & .314 & .506 & .292 & .416 & .263 & .495 & .473 \\
 & Direct MT & .184 & .359 & .346 & .362 & .316 & .338 & .377 & .322 & .298 & .370 & .299 & .255 & .359 & .220 & .322 & .267 & .311 & .216 & .394 & .370 \\
 & Task-Preserved & .405 & .548 & .518 & .514 & .494 & .485 & .508 & .482 & .494 & .527 & .436 & .388 & .540 & .281 & .512 & .372 & .433 & .290 & .543 & .540 \\
 & \datasettablee{} & \textbf{.512} & \textbf{.601} & \textbf{.554} & \textbf{.560} & \textbf{.506} & \textbf{.563} & \textbf{.575} & \textbf{.525} & \textbf{.571} & \textbf{.599} & \textbf{.508} & \textbf{.416} & \textbf{.562} & \textbf{.346} & \textbf{.543} & \textbf{.412} & \textbf{.525} & \textbf{.382} & \textbf{.584} & \textbf{.592} \\
\midrule
Phi-4-Mini-Instruct & Off-the-shelf & .312 & .282 & .294 & .316 & .277 & .202 & .277 & .333 & \textbf{.290} & .427 & .153 & \textbf{.231} & .325 & .164 & .357 & .216 & .135 & .122 & .329 & .252 \\
 & Direct MT & .115 & .181 & .224 & .229 & .227 & .158 & .220 & .255 & .185 & .266 & .153 & .118 & .255 & .126 & .216 & .174 & .153 & .115 & .227 & .214 \\
 & Task-Preserved & .252 & .279 & .285 & .320 & .261 & .200 & .316 & .375 & .238 & .409 & .163 & .189 & .323 & .182 & .388 & \textbf{.237} & \textbf{.179} & .153 & .322 & .269 \\
 & \datasettablee{} & \textbf{.354} & \textbf{.307} & \textbf{.323} & \textbf{.357} & \textbf{.322} & \textbf{.225} & \textbf{.346} & \textbf{.433} & .266 & \textbf{.508} & \textbf{.181} & .196 & \textbf{.338} & \textbf{.197} & \textbf{.431} & .231 & .174 & \textbf{.177} & \textbf{.335} & \textbf{.326} \\
\midrule
Ministral-3-3B-Instruct & Off-the-shelf & .258 & .309 & \textbf{.317} & .275 & \textbf{.305} & .214 & .336 & .298 & .248 & .333 & .251 & .205 & .316 & .164 & .274 & .193 & .176 & .139 & .264 & .300 \\
 & Direct MT & .117 & .191 & .195 & .231 & .181 & .170 & .203 & .214 & .211 & .253 & .165 & .168 & .209 & .172 & .166 & .197 & .172 & .158 & .255 & .198 \\
 & Task-Preserved & .270 & .284 & .283 & \textbf{.344} & .292 & .208 & .329 & .299 & .270 & .401 & .222 & \textbf{.227} & .311 & .266 & .331 & .218 & .219 & .177 & \textbf{.342} & .282 \\
 & \datasettablee{} & \textbf{.348} & \textbf{.324} & .279 & .338 & .283 & \textbf{.225} & \textbf{.348} & \textbf{.322} & \textbf{.320} & \textbf{.427} & \textbf{.255} & .216 & \textbf{.329} & \textbf{.293} & \textbf{.348} & \textbf{.281} & \textbf{.221} & \textbf{.189} & .336 & \textbf{.324} \\
\midrule
Llama-3.2-3B-Instruct & Off-the-shelf & .386 & .317 & \textbf{.382} & .299 & \textbf{.355} & .168 & \textbf{.392} & \textbf{.427} & .257 & \textbf{.523} & .200 & \textbf{.222} & \textbf{.318} & .222 & \textbf{.445} & .218 & .183 & .160 & \textbf{.409} & \textbf{.359} \\
 & Direct MT & .103 & .189 & .207 & .238 & .203 & .218 & .227 & .240 & .185 & .255 & .174 & .150 & .233 & .155 & .213 & .202 & .174 & .128 & .288 & .198 \\
 & Task-Preserved & .254 & .294 & .273 & .285 & .294 & .221 & .298 & .338 & .214 & .362 & .172 & .200 & .279 & .203 & .364 & \textbf{.256} & .189 & .174 & .338 & .277 \\
 & \datasettablee{} & \textbf{.421} & \textbf{.324} & .314 & \textbf{.373} & .349 & \textbf{.258} & .362 & \textbf{.427} & \textbf{.272} & .445 & \textbf{.226} & .220 & .311 & \textbf{.226} & .440 & .244 & \textbf{.233} & \textbf{.204} & .399 & .345 \\
\bottomrule
\end{tabular}%
}
\captionof{table}{\benchmark{} accuracy after language-specific fine-tuning (languages 1--20 of 50).}
\label{tab:ifeval-language-a}
\vspace{1.35\baselineskip}
\centering
\tiny
\setlength{\tabcolsep}{1.35pt}
\renewcommand{\arraystretch}{0.90}
\resizebox{0.98\textwidth}{!}{%
\begin{tabular}{llrrrrrrrrrrrrrrrrrrrr}
\toprule
\textbf{Model} & \textbf{Condition} & \textbf{hu} & \textbf{is} & \textbf{it} & \textbf{ka} & \textbf{lb} & \textbf{lij} & \textbf{lmo} & \textbf{lt} & \textbf{ltg} & \textbf{lv} & \textbf{mk} & \textbf{mt} & \textbf{nl} & \textbf{nno} & \textbf{nob} & \textbf{oc} & \textbf{pl} & \textbf{pt} & \textbf{ro} & \textbf{ru} \\
\midrule
Tiny-Aya-Global & Off-the-shelf & .442 & .288 & .457 & .189 & .235 & .365 & .332 & .485 & .265 & .479 & .377 & .498 & .530 & .559 & .576 & .368 & .418 & .436 & .433 & .409 \\
 & Direct MT & .262 & .227 & .355 & .202 & .233 & .179 & .246 & .294 & .183 & .284 & .278 & .359 & .327 & .319 & .389 & .284 & .255 & .338 & .290 & .272 \\
 & Task-Preserved & .427 & .271 & .567 & .233 & .332 & .338 & .384 & .523 & .256 & .416 & .429 & .494 & .549 & .475 & .521 & .443 & \textbf{.429} & .540 & .460 & \textbf{.510} \\
 & \datasettablee{} & \textbf{.484} & \textbf{.328} & \textbf{.584} & \textbf{.254} & \textbf{.382} & \textbf{.405} & \textbf{.435} & \textbf{.567} & \textbf{.357} & \textbf{.487} & \textbf{.525} & \textbf{.574} & \textbf{.553} & \textbf{.567} & \textbf{.605} & \textbf{.513} & .407 & \textbf{.571} & \textbf{.499} & .505 \\
\midrule
Phi-4-Mini-Instruct & Off-the-shelf & .238 & .193 & .336 & .206 & .191 & .202 & .216 & \textbf{.239} & .155 & .197 & \textbf{.314} & .200 & .298 & .296 & .338 & .275 & .251 & .355 & .268 & \textbf{.329} \\
 & Direct MT & .177 & .164 & .268 & .177 & .174 & .145 & .176 & .176 & .143 & .166 & .203 & .155 & .227 & .216 & .286 & .183 & .231 & .244 & .185 & .185 \\
 & Task-Preserved & .211 & .195 & .375 & \textbf{.231} & .187 & .181 & \textbf{.246} & .231 & .172 & .200 & .266 & \textbf{.229} & .335 & .296 & .376 & .254 & .281 & .372 & .242 & .292 \\
 & \datasettablee{} & \textbf{.253} & \textbf{.225} & \textbf{.401} & .221 & \textbf{.235} & \textbf{.239} & .235 & .227 & \textbf{.197} & \textbf{.244} & .291 & .223 & \textbf{.368} & \textbf{.344} & \textbf{.422} & \textbf{.309} & \textbf{.292} & \textbf{.429} & \textbf{.272} & \textbf{.329} \\
\midrule
Ministral-3-3B-Instruct & Off-the-shelf & .262 & .218 & .292 & .231 & .185 & .198 & .197 & .277 & .155 & \textbf{.252} & .284 & .172 & .261 & .303 & \textbf{.323} & .235 & .248 & .281 & .246 & .307 \\
 & Direct MT & .159 & .179 & .213 & .198 & .141 & .145 & .189 & .181 & .168 & .149 & .216 & .174 & .203 & .198 & .218 & .206 & .185 & .209 & .192 & .179 \\
 & Task-Preserved & .261 & .250 & .366 & .284 & .189 & .254 & .258 & .279 & .191 & .223 & .318 & .252 & \textbf{.331} & .313 & .321 & .282 & \textbf{.292} & \textbf{.364} & .290 & .336 \\
 & \datasettablee{} & \textbf{.277} & \textbf{.265} & \textbf{.386} & \textbf{.286} & \textbf{.223} & \textbf{.269} & \textbf{.311} & \textbf{.284} & \textbf{.208} & .239 & \textbf{.328} & \textbf{.256} & .320 & \textbf{.361} & \textbf{.323} & \textbf{.338} & .275 & \textbf{.364} & \textbf{.301} & \textbf{.362} \\
\midrule
Llama-3.2-3B-Instruct & Off-the-shelf & .290 & .221 & .449 & .177 & .229 & \textbf{.269} & .250 & \textbf{.219} & .158 & .195 & .316 & \textbf{.221} & .394 & .342 & \textbf{.397} & \textbf{.357} & .301 & .442 & \textbf{.364} & \textbf{.384} \\
 & Direct MT & .179 & .172 & .262 & .151 & .212 & .174 & .191 & .162 & .134 & .162 & .190 & .172 & .244 & .212 & .284 & .210 & .187 & .251 & .203 & .205 \\
 & Task-Preserved & .250 & .202 & .390 & .191 & .258 & .248 & .231 & .202 & .172 & .216 & .282 & .216 & .360 & .288 & .317 & .281 & .261 & .372 & .290 & .303 \\
 & \datasettablee{} & \textbf{.307} & \textbf{.256} & \textbf{.455} & \textbf{.225} & \textbf{.263} & .263 & \textbf{.298} & .210 & \textbf{.197} & \textbf{.218} & \textbf{.330} & \textbf{.221} & \textbf{.434} & \textbf{.366} & .389 & .342 & \textbf{.316} & \textbf{.451} & .323 & .351 \\
\bottomrule
\end{tabular}%
}
\captionof{table}{\benchmark{} accuracy after language-specific fine-tuning (languages 21--40 of 50).}
\label{tab:ifeval-language-b}
\vspace{1.35\baselineskip}

\centering
\tiny
\setlength{\tabcolsep}{1.35pt}
\renewcommand{\arraystretch}{0.90}
\resizebox{0.66\textwidth}{!}{%
\begin{tabular}{llrrrrrrrrrr}
\toprule
\textbf{Model} & \textbf{Condition} & \textbf{sc} & \textbf{scn} & \textbf{sk} & \textbf{sl} & \textbf{sq} & \textbf{sr} & \textbf{sv} & \textbf{szl} & \textbf{uk} & \textbf{vec} \\
\midrule
Tiny-Aya-Global & Off-the-shelf & .366 & .326 & .487 & .447 & .321 & .433 & .468 & .365 & .440 & .344 \\
 & Direct MT & .244 & .275 & .399 & .313 & .240 & .280 & .383 & .210 & .272 & .261 \\
 & Task-Preserved & .355 & .368 & .561 & .523 & .330 & .471 & .567 & .338 & .470 & .420 \\
 & \datasettablee{} & \textbf{.412} & \textbf{.441} & \textbf{.569} & \textbf{.552} & \textbf{.372} & \textbf{.594} & \textbf{.625} & \textbf{.393} & \textbf{.484} & \textbf{.466} \\
\midrule
Phi-4-Mini-Instruct & Off-the-shelf & .231 & .208 & .258 & .246 & .219 & \textbf{.295} & .338 & \textbf{.242} & .272 & .218 \\
 & Direct MT & .143 & .210 & .195 & .168 & .191 & .222 & .268 & .162 & .181 & .168 \\
 & Task-Preserved & .248 & .246 & .279 & .237 & \textbf{.246} & .289 & .349 & .214 & .264 & \textbf{.240} \\
 & \datasettablee{} & \textbf{.250} & \textbf{.252} & \textbf{.344} & \textbf{.254} & .237 & .291 & \textbf{.394} & .233 & \textbf{.279} & .237 \\
\midrule
Ministral-3-3B-Instruct & Off-the-shelf & .187 & .233 & \textbf{.303} & \textbf{.275} & .204 & .284 & \textbf{.342} & .202 & .299 & .202 \\
 & Direct MT & .160 & .166 & .197 & .164 & .187 & .215 & .209 & .166 & .155 & .183 \\
 & Task-Preserved & .227 & \textbf{.279} & .261 & .252 & .242 & .293 & .307 & .227 & .272 & .281 \\
 & \datasettablee{} & \textbf{.254} & .267 & .296 & \textbf{.275} & \textbf{.265} & \textbf{.307} & .314 & \textbf{.239} & \textbf{.305} & \textbf{.305} \\
\midrule
Llama-3.2-3B-Instruct & Off-the-shelf & .275 & .286 & \textbf{.380} & \textbf{.332} & \textbf{.254} & \textbf{.297} & \textbf{.434} & .227 & \textbf{.298} & .244 \\
 & Direct MT & .195 & .172 & .198 & .160 & .172 & .193 & .250 & .141 & .161 & .200 \\
 & Task-Preserved & .275 & .254 & .242 & .227 & .206 & .253 & .344 & .212 & .237 & .269 \\
 & \datasettablee{} & \textbf{.303} & \textbf{.303} & .309 & .269 & .242 & .295 & .375 & \textbf{.250} & .283 & \textbf{.284} \\
\bottomrule
\end{tabular}%
}
\captionof{table}{\benchmark{} accuracy  after language-specific fine-tuning (languages 41--50 of 50).}
\label{tab:ifeval-language-c}
\vspace{1.35\baselineskip}

\clearpage
\subsection{Aya Evaluation Suite: ROUGE-L after Language-Specific Fine-Tuning}
\vspace{7\baselineskip}
\centering
\tiny
\setlength{\tabcolsep}{1.35pt}
\renewcommand{\arraystretch}{0.90}
\resizebox{0.98\textwidth}{!}{%
\begin{tabular}{llrrrrrrrrrrrrrrrrrr}
\toprule
\textbf{Model} & \textbf{Condition} & \textbf{bg} & \textbf{ca} & \textbf{cs} & \textbf{cy} & \textbf{da} & \textbf{de} & \textbf{el} & \textbf{es} & \textbf{et} & \textbf{eu} & \textbf{fi} & \textbf{fr} & \textbf{ga} & \textbf{gd} & \textbf{gl} & \textbf{hu} & \textbf{is} & \textbf{it} \\
\midrule
Tiny-Aya-Global & Off-the-shelf & .041 & .099 & .134 & .116 & .134 & .106 & .053 & .095 & .116 & .129 & .118 & .116 & .119 & .083 & .113 & .131 & .087 & .109 \\
 & Direct MT & \textbf{.123} & .271 & .248 & .235 & .272 & \textbf{.258} & .158 & .294 & .222 & .176 & .208 & .279 & .236 & .198 & .289 & .269 & .203 & .275 \\
 & Task-Preserved & .114 & .283 & \textbf{.265} & .254 & \textbf{.284} & \textbf{.258} & .157 & .299 & \textbf{.232} & .184 & \textbf{.219} & \textbf{.290} & .248 & .188 & \textbf{.299} & .278 & .214 & .283 \\
 & \datasettablee{} & .122 & \textbf{.285} & .255 & \textbf{.259} & .280 & \textbf{.258} & \textbf{.160} & \textbf{.312} & \textbf{.232} & \textbf{.188} & .212 & .285 & \textbf{.251} & \textbf{.216} & .294 & \textbf{.287} & \textbf{.220} & \textbf{.288} \\
\midrule
Phi-4-Mini-Instruct & Off-the-shelf & .073 & .123 & .106 & .072 & .103 & .078 & .087 & .158 & .045 & .052 & .078 & .129 & .073 & .066 & .091 & .106 & .057 & .130 \\
 & Direct MT & \textbf{.126} & .259 & .217 & .160 & .261 & .258 & .130 & .294 & .147 & .127 & .184 & .267 & .167 & .155 & .276 & .235 & .173 & .245 \\
 & Task-Preserved & .109 & .251 & .218 & .166 & .259 & .255 & .138 & .287 & .144 & .130 & .178 & .273 & .170 & .165 & .276 & .234 & .176 & .255 \\
 & \datasettablee{} & .117 & \textbf{.276} & \textbf{.241} & \textbf{.200} & \textbf{.267} & \textbf{.267} & \textbf{.157} & \textbf{.300} & \textbf{.172} & \textbf{.148} & \textbf{.200} & \textbf{.282} & \textbf{.195} & \textbf{.203} & \textbf{.293} & \textbf{.261} & \textbf{.200} & \textbf{.267} \\
\midrule
Ministral-3-3B-Instruct & Off-the-shelf & .056 & .163 & .138 & .105 & .156 & .117 & .066 & .158 & .100 & .063 & .115 & .161 & .098 & .092 & .166 & .135 & .129 & .140 \\
 & Direct MT & .095 & .255 & .228 & .166 & .245 & .218 & .140 & .278 & .172 & .127 & .183 & .252 & .148 & .149 & .261 & .210 & .181 & .252 \\
 & Task-Preserved & .099 & .256 & .221 & .165 & .241 & .213 & .132 & .279 & .174 & .136 & .181 & .256 & .163 & .149 & .265 & .199 & .192 & .258 \\
 & \datasettablee{} & \textbf{.100} & \textbf{.272} & \textbf{.254} & \textbf{.214} & \textbf{.272} & \textbf{.246} & \textbf{.153} & \textbf{.287} & \textbf{.199} & \textbf{.154} & \textbf{.203} & \textbf{.286} & \textbf{.197} & \textbf{.186} & \textbf{.281} & \textbf{.239} & \textbf{.221} & \textbf{.273} \\
\midrule
Llama-3.2-3B-Instruct & Off-the-shelf & .090 & .222 & .220 & .138 & .234 & .205 & .133 & .254 & .163 & .125 & .163 & .247 & .132 & .159 & .229 & .225 & .194 & .243 \\
 & Direct MT & .078 & .254 & .219 & .172 & .243 & .230 & .121 & .285 & .166 & .141 & .170 & .266 & .166 & .170 & .271 & .234 & .185 & .261 \\
 & Task-Preserved & .085 & .267 & .217 & .175 & .236 & .250 & \textbf{.149} & .296 & .175 & .131 & .178 & \textbf{.273} & .166 & .167 & .274 & .238 & .191 & .287 \\
 & \datasettablee{} & \textbf{.112} & \textbf{.283} & \textbf{.241} & \textbf{.203} & \textbf{.271} & \textbf{.254} & .133 & \textbf{.302} & \textbf{.193} & \textbf{.164} & \textbf{.183} & .267 & \textbf{.188} & \textbf{.215} & \textbf{.291} & \textbf{.264} & \textbf{.202} & \textbf{.288} \\
\bottomrule
\end{tabular}%
}
\captionof{table}{ROUGE-L on Aya Evaluation Suite after language-specific fine-tuning (languages 1--18 of 36).}
\label{tab:rouge-language-a}
\vspace{8\baselineskip}
\centering
\tiny
\setlength{\tabcolsep}{1.35pt}
\renewcommand{\arraystretch}{0.90}
\resizebox{0.98\textwidth}{!}{%
\begin{tabular}{llrrrrrrrrrrrrrrrrrr}
\toprule
\textbf{Model} & \textbf{Condition} & \textbf{lb} & \textbf{lt} & \textbf{lv} & \textbf{mk} & \textbf{mt} & \textbf{nl} & \textbf{nno} & \textbf{nob} & \textbf{pl} & \textbf{pt} & \textbf{ro} & \textbf{ru} & \textbf{sk} & \textbf{sl} & \textbf{sq} & \textbf{sr} & \textbf{sv} & \textbf{uk} \\
\midrule
Tiny-Aya-Global & Off-the-shelf & .047 & .089 & .083 & .047 & .120 & .134 & .162 & .171 & .106 & .105 & .123 & .047 & .119 & .118 & .128 & .033 & .099 & .085 \\
 & Direct MT & .205 & .206 & .215 & .125 & .231 & .279 & .238 & .284 & .228 & .305 & .263 & .131 & .247 & .243 & .220 & .130 & \textbf{.284} & \textbf{.157} \\
 & Task-Preserved & .184 & .212 & .221 & .120 & .231 & .278 & .240 & .282 & \textbf{.242} & .309 & \textbf{.282} & .127 & .250 & .249 & .231 & .134 & .274 & .140 \\
 & \datasettablee{} & \textbf{.221} & \textbf{.213} & \textbf{.223} & \textbf{.134} & \textbf{.260} & \textbf{.286} & \textbf{.245} & \textbf{.291} & .238 & \textbf{.311} & .273 & \textbf{.147} & \textbf{.264} & \textbf{.255} & \textbf{.251} & \textbf{.135} & .281 & .148 \\
\midrule
Phi-4-Mini-Instruct & Off-the-shelf & .059 & .088 & .089 & .083 & .047 & .089 & .067 & .102 & .114 & .143 & .121 & .091 & .079 & .063 & .116 & .072 & .126 & .100 \\
 & Direct MT & .190 & .163 & .166 & .124 & .140 & .262 & .204 & .259 & .216 & .286 & .227 & .117 & .203 & .189 & .204 & .122 & .251 & .139 \\
 & Task-Preserved & .192 & .170 & .176 & .129 & .151 & .253 & .206 & .259 & .222 & .292 & .235 & .119 & .193 & .195 & .219 & .120 & .253 & .136 \\
 & \datasettablee{} & \textbf{.227} & \textbf{.195} & \textbf{.184} & \textbf{.137} & \textbf{.169} & \textbf{.276} & \textbf{.236} & \textbf{.283} & \textbf{.227} & \textbf{.310} & \textbf{.250} & \textbf{.128} & \textbf{.216} & \textbf{.211} & \textbf{.233} & \textbf{.144} & \textbf{.271} & \textbf{.141} \\
\midrule
Ministral-3-3B-Instruct & Off-the-shelf & .078 & .090 & .103 & .071 & .066 & .156 & .147 & .158 & .122 & .172 & .149 & .071 & .130 & .116 & .132 & .091 & .171 & .052 \\
 & Direct MT & .164 & .168 & .165 & .102 & .157 & .236 & .216 & .253 & .203 & .273 & .230 & .095 & .214 & .175 & .211 & .100 & .244 & .113 \\
 & Task-Preserved & .155 & .174 & .164 & .115 & .162 & .243 & .213 & .244 & .208 & .279 & .222 & .093 & .207 & .195 & .208 & .103 & .259 & \textbf{.133} \\
 & \datasettablee{} & \textbf{.209} & \textbf{.191} & \textbf{.195} & \textbf{.126} & \textbf{.202} & \textbf{.263} & \textbf{.239} & \textbf{.263} & \textbf{.224} & \textbf{.298} & \textbf{.257} & \textbf{.113} & \textbf{.227} & \textbf{.212} & \textbf{.239} & \textbf{.117} & \textbf{.276} & .132 \\
\midrule
Llama-3.2-3B-Instruct & Off-the-shelf & .165 & .176 & \textbf{.184} & .098 & .141 & .257 & .203 & .253 & .211 & .274 & .250 & .093 & .179 & .188 & .178 & .079 & \textbf{.275} & .110 \\
 & Direct MT & .196 & .157 & .165 & .111 & .163 & .270 & .220 & .256 & .204 & .286 & .226 & .083 & .202 & .188 & .215 & .098 & .256 & .124 \\
 & Task-Preserved & .221 & .181 & .174 & .090 & .174 & .262 & .221 & .251 & .206 & .285 & .241 & \textbf{.107} & .190 & .185 & .225 & .090 & .258 & .123 \\
 & \datasettablee{} & \textbf{.227} & \textbf{.202} & .182 & \textbf{.120} & \textbf{.199} & \textbf{.281} & \textbf{.233} & \textbf{.267} & \textbf{.222} & \textbf{.303} & \textbf{.264} & .086 & \textbf{.210} & \textbf{.210} & \textbf{.257} & \textbf{.121} & \textbf{.275} & \textbf{.127} \\
\bottomrule
\end{tabular}%
}
\captionof{table}{ROUGE-L on Aya Evaluation Suite after language-specific fine-tuning (languages 19--36 of 36).}
\label{tab:rouge-language-b}
\vspace{0.4\baselineskip}

\clearpage
\subsection{Aya Evaluation Suite: F\textsc{bert} after Language-Specific Fine-Tuning}
\vspace{7\baselineskip}
\centering
\tiny
\setlength{\tabcolsep}{1.35pt}
\renewcommand{\arraystretch}{0.90}
\resizebox{0.98\textwidth}{!}{%
\begin{tabular}{llrrrrrrrrrrrrrrrrrr}
\toprule
\textbf{Model} & \textbf{Condition} & \textbf{bg} & \textbf{ca} & \textbf{cs} & \textbf{cy} & \textbf{da} & \textbf{de} & \textbf{el} & \textbf{es} & \textbf{et} & \textbf{eu} & \textbf{fi} & \textbf{fr} & \textbf{ga} & \textbf{gd} & \textbf{gl} & \textbf{hu} & \textbf{is} & \textbf{it} \\
\midrule
Tiny-Aya-Global & Off-the-shelf & .822 & .822 & .840 & .825 & .834 & .826 & .826 & .819 & .832 & .821 & .827 & .829 & .821 & .791 & .829 & .831 & .798 & .828 \\
 & Direct MT & .881 & .876 & .878 & .860 & .883 & .884 & .881 & .885 & .873 & .862 & .873 & .884 & .854 & .828 & \textbf{.885} & .877 & .845 & .883 \\
 & Task-Preserved & .884 & .881 & \textbf{.882} & .865 & .886 & .884 & .883 & .887 & .874 & .866 & .877 & \textbf{.887} & .856 & .824 & \textbf{.885} & .880 & .848 & .886 \\
 & \datasettablee{} & \textbf{.886} & \textbf{.883} & \textbf{.882} & \textbf{.867} & \textbf{.887} & \textbf{.886} & \textbf{.887} & \textbf{.890} & \textbf{.877} & \textbf{.869} & \textbf{.878} & \textbf{.887} & \textbf{.857} & \textbf{.835} & \textbf{.885} & \textbf{.885} & \textbf{.855} & \textbf{.887} \\
\midrule
Phi-4-Mini-Instruct & Off-the-shelf & .839 & .842 & .838 & .797 & .842 & .841 & .835 & .858 & .796 & .800 & .830 & .853 & .789 & .772 & .833 & .831 & .794 & .848 \\
 & Direct MT & .870 & .870 & .864 & .819 & .876 & .881 & .853 & .884 & .840 & .828 & .859 & .880 & .820 & .805 & .876 & .863 & .833 & .875 \\
 & Task-Preserved & .872 & .866 & .868 & .824 & .878 & .882 & .862 & .884 & .835 & .829 & .862 & .882 & .820 & .814 & .874 & .862 & .840 & .877 \\
 & \datasettablee{} & \textbf{.876} & \textbf{.878} & \textbf{.874} & \textbf{.840} & \textbf{.884} & \textbf{.885} & \textbf{.867} & \textbf{.888} & \textbf{.850} & \textbf{.839} & \textbf{.872} & \textbf{.885} & \textbf{.832} & \textbf{.827} & \textbf{.879} & \textbf{.872} & \textbf{.845} & \textbf{.884} \\
\midrule
Ministral-3-3B-Instruct & Off-the-shelf & .841 & .843 & .839 & .803 & .843 & .837 & .829 & .846 & .821 & .806 & .837 & .847 & .790 & .774 & .845 & .824 & .815 & .842 \\
 & Direct MT & .863 & .862 & .863 & .816 & .869 & .867 & .857 & .873 & .840 & .829 & .849 & .873 & .799 & .797 & .870 & .850 & .828 & .872 \\
 & Task-Preserved & .865 & .868 & .864 & .815 & .869 & .870 & .864 & .874 & .845 & .838 & .856 & .880 & .813 & .800 & .875 & .845 & .839 & .877 \\
 & \datasettablee{} & \textbf{.878} & \textbf{.876} & \textbf{.876} & \textbf{.842} & \textbf{.882} & \textbf{.882} & \textbf{.879} & \textbf{.885} & \textbf{.859} & \textbf{.849} & \textbf{.872} & \textbf{.886} & \textbf{.831} & \textbf{.822} & \textbf{.878} & \textbf{.861} & \textbf{.852} & \textbf{.883} \\
\midrule
Llama-3.2-3B-Instruct & Off-the-shelf & .864 & .868 & .870 & .826 & .875 & .871 & .864 & .878 & .856 & .850 & .861 & .878 & .827 & .823 & .866 & .867 & .846 & .879 \\
 & Direct MT & .855 & .869 & .861 & .829 & .876 & .874 & .861 & .883 & .845 & .837 & .854 & .879 & .820 & .820 & .876 & .858 & .836 & .878 \\
 & Task-Preserved & .859 & .874 & .863 & .830 & .873 & .881 & .864 & .885 & .850 & .838 & .861 & \textbf{.883} & .822 & .816 & .878 & .863 & .839 & .885 \\
 & \datasettablee{} & \textbf{.875} & \textbf{.881} & \textbf{.873} & \textbf{.848} & \textbf{.880} & \textbf{.884} & \textbf{.877} & \textbf{.890} & \textbf{.861} & \textbf{.851} & \textbf{.869} & .882 & \textbf{.833} & \textbf{.839} & \textbf{.882} & \textbf{.876} & \textbf{.849} & \textbf{.888} \\
\bottomrule
\end{tabular}%
}
\captionof{table}{F\textsc{bert} on Aya Evaluation Suite after language-specific fine-tuning (languages 1--18 of 36).}
\label{tab:bert-language-a}
\vspace{7\baselineskip}
\centering
\tiny
\setlength{\tabcolsep}{1.35pt}
\renewcommand{\arraystretch}{0.90}
\resizebox{0.98\textwidth}{!}{%
\begin{tabular}{llrrrrrrrrrrrrrrrrrr}
\toprule
\textbf{Model} & \textbf{Condition} & \textbf{lb} & \textbf{lt} & \textbf{lv} & \textbf{mk} & \textbf{mt} & \textbf{nl} & \textbf{nno} & \textbf{nob} & \textbf{pl} & \textbf{pt} & \textbf{ro} & \textbf{ru} & \textbf{sk} & \textbf{sl} & \textbf{sq} & \textbf{sr} & \textbf{sv} & \textbf{uk} \\
\midrule
Tiny-Aya-Global & Off-the-shelf & .771 & .814 & .810 & .817 & .818 & .835 & .849 & .856 & .826 & .824 & .832 & .828 & .835 & .824 & .818 & .815 & .818 & .841 \\
 & Direct MT & .826 & .873 & .871 & .875 & .852 & .883 & .870 & .888 & .876 & .887 & .879 & .883 & .877 & .879 & .855 & .876 & .884 & .877 \\
 & Task-Preserved & .820 & .873 & .873 & .874 & .852 & .885 & .872 & .889 & .881 & .888 & \textbf{.883} & .885 & .878 & .881 & .857 & .876 & .885 & .882 \\
 & \datasettablee{} & \textbf{.835} & \textbf{.875} & \textbf{.874} & \textbf{.878} & \textbf{.858} & \textbf{.887} & \textbf{.874} & \textbf{.890} & \textbf{.882} & \textbf{.889} & .882 & \textbf{.889} & \textbf{.885} & \textbf{.884} & \textbf{.868} & \textbf{.881} & \textbf{.887} & \textbf{.884} \\
\midrule
Phi-4-Mini-Instruct & Off-the-shelf & .783 & .824 & .817 & .837 & .781 & .838 & .818 & .841 & .842 & .855 & .836 & .857 & .828 & .818 & .825 & .830 & .843 & .847 \\
 & Direct MT & .819 & .846 & .843 & .867 & .819 & .878 & .858 & .877 & .866 & .883 & .861 & .883 & .859 & .857 & .846 & .864 & .872 & .870 \\
 & Task-Preserved & .821 & .850 & .848 & .865 & .819 & .877 & .859 & .876 & .870 & .883 & .866 & .885 & .860 & .855 & .853 & .864 & .875 & .873 \\
 & \datasettablee{} & \textbf{.832} & \textbf{.857} & \textbf{.853} & \textbf{.872} & \textbf{.830} & \textbf{.883} & \textbf{.867} & \textbf{.886} & \textbf{.876} & \textbf{.888} & \textbf{.872} & \textbf{.887} & \textbf{.867} & \textbf{.865} & \textbf{.858} & \textbf{.873} & \textbf{.882} & \textbf{.878} \\
\midrule
Ministral-3-3B-Instruct & Off-the-shelf & .790 & .815 & .816 & .837 & .784 & .845 & .837 & .846 & .839 & .848 & .839 & .846 & .832 & .829 & .822 & .838 & .847 & .834 \\
 & Direct MT & .809 & .841 & .839 & .854 & .819 & .864 & .855 & .872 & .865 & .876 & .859 & .869 & .857 & .847 & .844 & .862 & .867 & .857 \\
 & Task-Preserved & .808 & .847 & .839 & .857 & .820 & .868 & .860 & .873 & .872 & .879 & .860 & .877 & .860 & .854 & .844 & .858 & .869 & .864 \\
 & \datasettablee{} & \textbf{.826} & \textbf{.862} & \textbf{.856} & \textbf{.872} & \textbf{.839} & \textbf{.879} & \textbf{.868} & \textbf{.880} & \textbf{.875} & \textbf{.884} & \textbf{.873} & \textbf{.884} & \textbf{.868} & \textbf{.870} & \textbf{.860} & \textbf{.872} & \textbf{.882} & \textbf{.875} \\
\midrule
Llama-3.2-3B-Instruct & Off-the-shelf & .815 & .859 & \textbf{.859} & .857 & .823 & .879 & .865 & .878 & .872 & .879 & .874 & .878 & .860 & .860 & .845 & .852 & .878 & .870 \\
 & Direct MT & .823 & .849 & .849 & .858 & .826 & .878 & .864 & .875 & .866 & .883 & .864 & .870 & .857 & .857 & .847 & .861 & .871 & .864 \\
 & Task-Preserved & .830 & .856 & .852 & .865 & .830 & .879 & .864 & .874 & .868 & .884 & .870 & .875 & .855 & .856 & .854 & .862 & .869 & .868 \\
 & \datasettablee{} & \textbf{.837} & \textbf{.867} & .858 & \textbf{.872} & \textbf{.840} & \textbf{.884} & \textbf{.869} & \textbf{.880} & \textbf{.877} & \textbf{.889} & \textbf{.881} & \textbf{.881} & \textbf{.866} & \textbf{.869} & \textbf{.867} & \textbf{.869} & \textbf{.879} & \textbf{.878} \\
\bottomrule
\end{tabular}%
}
\captionof{table}{F\textsc{bert} on Aya Evaluation Suite after language-specific fine-tuning (languages 19--36 of 36).}
\label{tab:bert-language-b}
\vspace{0.4\baselineskip}

\clearpage
\subsection{Per-Language Primary Comparison}
\label{per_lang_comparison}
\vspace{3\baselineskip}
\centering
\includegraphics[width=\textwidth]{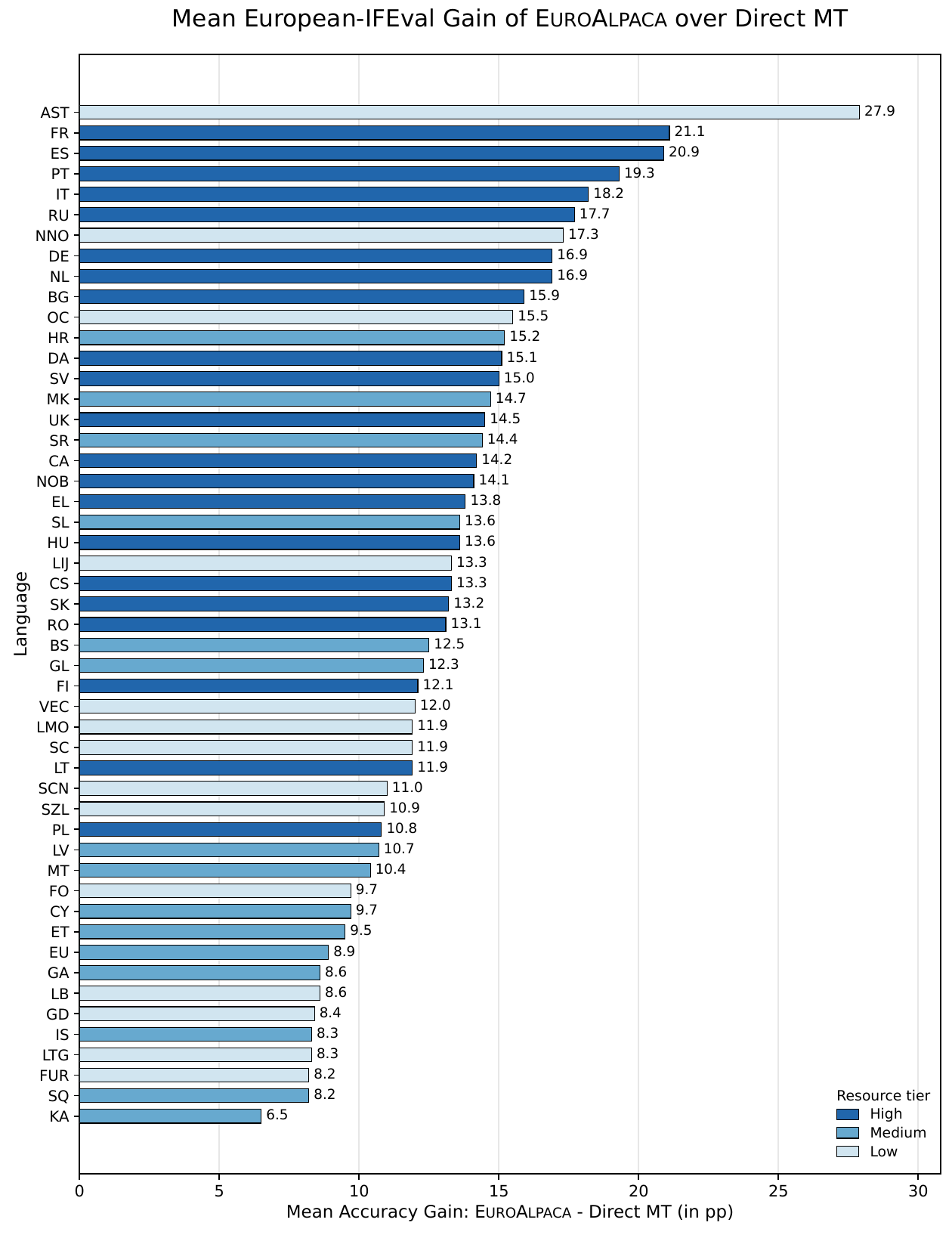}
\captionof{figure}{Mean \benchmark{} accuracy gain of \loc{} over direct MT for each language, averaged across the four evaluated models. Languages are ordered by the magnitude of the gain and coloured by their internal resource tier. All 50 language-level gains are positive.}
\label{fig:ifeval-language-grouped}
\vspace{0.5em}

\clearpage
% Pairwise tables are placed as ordinary, non-floating material.
% Do not force a new page between resources: LaTeX will use the remaining
% space and move a whole resizebox to the following page only when needed.
\subsection{Pairwise Comparisons with Existing Alpaca-Derived Resources on \benchmarktitle{} Accuracy}
\label{app:external-pairwise-results}

\subsubsection{Okapi}
\nopagebreak[4]
\vspace{1.35\baselineskip}
\centering
\tiny
\setlength{\tabcolsep}{1.35pt}
\renewcommand{\arraystretch}{0.90}
\resizebox{0.98\textwidth}{!}{%
\begin{tabular}{llrrrrrrrrrrrrrrrrr}
\toprule
\textbf{Model} & \textbf{Training resource} & \textbf{ca} & \textbf{da} & \textbf{de} & \textbf{es} & \textbf{eu} & \textbf{fr} & \textbf{hr} & \textbf{hu} & \textbf{it} & \textbf{nl} & \textbf{pt} & \textbf{ro} & \textbf{ru} & \textbf{sk} & \textbf{sr} & \textbf{sv} & \textbf{uk} \\
\midrule
Tiny-Aya-Global & Okapi & \textbf{.573} & .566 & \textbf{.529} & .573 & .410 & .481 & .582 & .468 & .492 & .508 & .490 & .449 & .471 & \textbf{.605} & .579 & .555 & .460 \\
 & \datasettablee{} & .560 & \textbf{.575} & .525 & \textbf{.599} & \textbf{.416} & \textbf{.543} & \textbf{.592} & \textbf{.484} & \textbf{.584} & \textbf{.553} & \textbf{.571} & \textbf{.499} & \textbf{.505} & .569 & \textbf{.594} & \textbf{.625} & \textbf{.484} \\
\midrule
Phi-4-Mini-Instruct & Okapi & .348 & \textbf{.357} & \textbf{.442} & .449 & \textbf{.207} & .423 & .305 & \textbf{.266} & .399 & \textbf{.384} & .360 & \textbf{.275} & \textbf{.342} & .323 & \textbf{.318} & \textbf{.399} & \textbf{.336} \\
 & \datasettablee{} & \textbf{.357} & .346 & .433 & \textbf{.508} & .196 & \textbf{.431} & \textbf{.326} & .253 & \textbf{.401} & .368 & \textbf{.429} & .272 & .329 & \textbf{.344} & .291 & .394 & .279 \\
\midrule
Ministral-3-3B-Instruct & Okapi & \textbf{.351} & \textbf{.372} & \textbf{.372} & .392 & .213 & .338 & \textbf{.330} & .257 & .368 & .312 & .331 & .294 & .322 & .288 & \textbf{.330} & \textbf{.314} & .301 \\
 & \datasettablee{} & .338 & .348 & .322 & \textbf{.427} & \textbf{.216} & \textbf{.348} & .324 & \textbf{.277} & \textbf{.386} & \textbf{.320} & \textbf{.364} & \textbf{.301} & \textbf{.362} & \textbf{.296} & .307 & \textbf{.314} & \textbf{.305} \\
\midrule
Llama-3.2-3B-Instruct & Okapi & .320 & .311 & .403 & .375 & .216 & .364 & .315 & .246 & .390 & .368 & .377 & .294 & .296 & .298 & .276 & .360 & .218 \\
 & \datasettablee{} & \textbf{.373} & \textbf{.362} & \textbf{.427} & \textbf{.445} & \textbf{.220} & \textbf{.440} & \textbf{.345} & \textbf{.307} & \textbf{.455} & \textbf{.434} & \textbf{.451} & \textbf{.323} & \textbf{.351} & \textbf{.309} & \textbf{.295} & \textbf{.375} & \textbf{.283} \\
\bottomrule
\end{tabular}%
}
\nopagebreak[4]
\captionof{table}{Pairwise comparison of Okapi and \dataset{} after language-specific fine-tuning on \benchmark{} over their shared-language subset.}
\label{tab:okapi-pairwise}

\par\medskip
\subsubsection{MITS}
\nopagebreak[4]
\vspace{1.35\baselineskip}
\centering
\tiny
\setlength{\tabcolsep}{1.35pt}
\renewcommand{\arraystretch}{0.90}
\resizebox{0.98\textwidth}{!}{%
\begin{tabular}{llrrrrrrrrrrrrrrrrrrr}
\toprule
\textbf{Model} & \textbf{Training resource} & \textbf{bg} & \textbf{bs} & \textbf{ca} & \textbf{cs} & \textbf{cy} & \textbf{da} & \textbf{de} & \textbf{el} & \textbf{es} & \textbf{et} & \textbf{eu} & \textbf{fi} & \textbf{fr} & \textbf{ga} & \textbf{gd} & \textbf{gl} & \textbf{hr} & \textbf{hu} & \textbf{is} \\
\midrule
Tiny-Aya-Global & MITS & .489 & .468 & .453 & .384 & .473 & .477 & .429 & .455 & .458 & .433 & .360 & .512 & .442 & .435 & .267 & .479 & .517 & .364 & .269 \\
 & \datasettablee{} & \textbf{.601} & \textbf{.554} & \textbf{.560} & \textbf{.506} & \textbf{.563} & \textbf{.575} & \textbf{.525} & \textbf{.571} & \textbf{.599} & \textbf{.508} & \textbf{.416} & \textbf{.562} & \textbf{.543} & \textbf{.525} & \textbf{.382} & \textbf{.584} & \textbf{.592} & \textbf{.484} & \textbf{.328} \\
\midrule
Phi-4-Mini-Instruct & MITS & .296 & .312 & .296 & .294 & .162 & .314 & .316 & .237 & .373 & .159 & .187 & .305 & .359 & .166 & .135 & .307 & .281 & .227 & .189 \\
 & \datasettablee{} & \textbf{.307} & \textbf{.323} & \textbf{.357} & \textbf{.322} & \textbf{.225} & \textbf{.346} & \textbf{.433} & \textbf{.266} & \textbf{.508} & \textbf{.181} & \textbf{.196} & \textbf{.338} & \textbf{.431} & \textbf{.174} & \textbf{.177} & \textbf{.335} & \textbf{.326} & \textbf{.253} & \textbf{.225} \\
\midrule
Ministral-3-3B-Instruct & MITS & .244 & .228 & .264 & .196 & .158 & .264 & .244 & .244 & .335 & .207 & .172 & .272 & .253 & .174 & \textbf{.189} & .274 & .235 & .200 & .223 \\
 & \datasettablee{} & \textbf{.324} & \textbf{.279} & \textbf{.338} & \textbf{.283} & \textbf{.225} & \textbf{.348} & \textbf{.322} & \textbf{.320} & \textbf{.427} & \textbf{.255} & \textbf{.216} & \textbf{.329} & \textbf{.348} & \textbf{.221} & \textbf{.189} & \textbf{.336} & \textbf{.324} & \textbf{.277} & \textbf{.265} \\
\midrule
Llama-3.2-3B-Instruct & MITS & .231 & .289 & .274 & .251 & .185 & .303 & .281 & .235 & .346 & .190 & .198 & .285 & .364 & .160 & .170 & .351 & .254 & .226 & .202 \\
 & \datasettablee{} & \textbf{.324} & \textbf{.314} & \textbf{.373} & \textbf{.349} & \textbf{.258} & \textbf{.362} & \textbf{.427} & \textbf{.272} & \textbf{.445} & \textbf{.226} & \textbf{.220} & \textbf{.311} & \textbf{.440} & \textbf{.233} & \textbf{.204} & \textbf{.399} & \textbf{.345} & \textbf{.307} & \textbf{.256} \\
\bottomrule
\end{tabular}%
}
\nopagebreak[4]
\captionof{table}{Pairwise comparison of MITS and \dataset{} after language-specific fine-tuning on \benchmark{} accuracy over their shared-language subset (languages 1--19 of 38).}
\label{tab:mits-pairwise-a}

\vspace{1.35\baselineskip}
\centering
\tiny
\setlength{\tabcolsep}{1.35pt}
\renewcommand{\arraystretch}{0.90}
\resizebox{0.98\textwidth}{!}{%
\begin{tabular}{llrrrrrrrrrrrrrrrrrrr}
\toprule
\textbf{Model} & \textbf{Training resource} & \textbf{it} & \textbf{ka} & \textbf{lb} & \textbf{lt} & \textbf{lv} & \textbf{mk} & \textbf{mt} & \textbf{nl} & \textbf{nob} & \textbf{pl} & \textbf{pt} & \textbf{ro} & \textbf{ru} & \textbf{sk} & \textbf{sl} & \textbf{sq} & \textbf{sr} & \textbf{sv} & \textbf{uk} \\
\midrule
Tiny-Aya-Global & MITS & .482 & .235 & .273 & .429 & .408 & .375 & .460 & .444 & .552 & .298 & .449 & .373 & .342 & .506 & .473 & .275 & .381 & .545 & .355 \\
 & \datasettablee{} & \textbf{.584} & \textbf{.254} & \textbf{.382} & \textbf{.567} & \textbf{.487} & \textbf{.525} & \textbf{.574} & \textbf{.553} & \textbf{.605} & \textbf{.407} & \textbf{.571} & \textbf{.499} & \textbf{.505} & \textbf{.569} & \textbf{.552} & \textbf{.372} & \textbf{.594} & \textbf{.625} & \textbf{.484} \\
\midrule
Phi-4-Mini-Instruct & MITS & .370 & .193 & .191 & .216 & .176 & .289 & .191 & .314 & .366 & .264 & .333 & .211 & .266 & .273 & \textbf{.260} & .208 & .249 & .364 & .261 \\
 & \datasettablee{} & \textbf{.401} & \textbf{.221} & \textbf{.235} & \textbf{.227} & \textbf{.244} & \textbf{.291} & \textbf{.223} & \textbf{.368} & \textbf{.422} & \textbf{.292} & \textbf{.429} & \textbf{.272} & \textbf{.329} & \textbf{.344} & .254 & \textbf{.237} & \textbf{.291} & \textbf{.394} & \textbf{.279} \\
\midrule
Ministral-3-3B-Instruct & MITS & .279 & .218 & .219 & .206 & .221 & .234 & .233 & .268 & .261 & .183 & .296 & .200 & .213 & .233 & .216 & .195 & .211 & .292 & .203 \\
 & \datasettablee{} & \textbf{.386} & \textbf{.286} & \textbf{.223} & \textbf{.284} & \textbf{.239} & \textbf{.328} & \textbf{.256} & \textbf{.320} & \textbf{.323} & \textbf{.275} & \textbf{.364} & \textbf{.301} & \textbf{.362} & \textbf{.296} & \textbf{.275} & \textbf{.265} & \textbf{.307} & \textbf{.314} & \textbf{.305} \\
\midrule
Llama-3.2-3B-Instruct & MITS & .383 & .189 & .191 & .185 & .181 & .238 & .197 & .336 & .340 & .226 & .344 & .266 & .231 & .239 & .239 & .179 & .218 & .320 & .207 \\
 & \datasettablee{} & \textbf{.455} & \textbf{.225} & \textbf{.263} & \textbf{.210} & \textbf{.218} & \textbf{.330} & \textbf{.221} & \textbf{.434} & \textbf{.389} & \textbf{.316} & \textbf{.451} & \textbf{.323} & \textbf{.351} & \textbf{.309} & \textbf{.269} & \textbf{.242} & \textbf{.295} & \textbf{.375} & \textbf{.283} \\
\bottomrule
\end{tabular}%
}
\nopagebreak[4]
\captionof{table}{Pairwise comparison of MITS and \dataset{} after language-specific fine-tuning on \benchmark{} accuracy over their shared-language subset (languages 20--38 of 38).}
\label{tab:mits-pairwise-b}

\par\medskip
\subsubsection{Bactrian-X}
\nopagebreak[4]
\vspace{1.35\baselineskip}
\centering
\tiny
\setlength{\tabcolsep}{1.35pt}
\renewcommand{\arraystretch}{0.90}
\resizebox{0.98\textwidth}{!}{%
\begin{tabular}{llrrrrrrrrrrrrrrrrrrrr}
\toprule
\textbf{Model} & \textbf{Training resource} & \textbf{cs} & \textbf{de} & \textbf{es} & \textbf{et} & \textbf{fi} & \textbf{fr} & \textbf{gl} & \textbf{hr} & \textbf{it} & \textbf{lt} & \textbf{lv} & \textbf{mk} & \textbf{nl} & \textbf{pl} & \textbf{pt} & \textbf{ro} & \textbf{ru} & \textbf{sl} & \textbf{sv} & \textbf{uk} \\
\midrule
Tiny-Aya-Global & Bactrian-X & .466 & .460 & \textbf{.612} & .490 & \textbf{.577} & \textbf{.586} & \textbf{.608} & .576 & \textbf{.614} & .557 & \textbf{.510} & .506 & \textbf{.577} & \textbf{.481} & \textbf{.608} & .429 & \textbf{.571} & \textbf{.584} & \textbf{.640} & \textbf{.545} \\
 & \datasettablee{} & \textbf{.506} & \textbf{.525} & .599 & \textbf{.508} & .562 & .543 & .584 & \textbf{.592} & .584 & \textbf{.567} & .487 & \textbf{.525} & .553 & .407 & .571 & \textbf{.499} & .505 & .552 & .625 & .484 \\
\midrule
Phi-4-Mini-Instruct & Bactrian-X & \textbf{.329} & \textbf{.447} & .468 & .170 & \textbf{.366} & \textbf{.473} & \textbf{.410} & \textbf{.328} & \textbf{.458} & \textbf{.242} & .225 & \textbf{.364} & \textbf{.423} & \textbf{.312} & \textbf{.471} & \textbf{.290} & \textbf{.377} & \textbf{.254} & \textbf{.416} & \textbf{.336} \\
 & \datasettablee{} & .322 & .433 & \textbf{.508} & \textbf{.181} & .338 & .431 & .335 & .326 & .401 & .227 & \textbf{.244} & .291 & .368 & .292 & .429 & .272 & .329 & \textbf{.254} & .394 & .279 \\
\midrule
Ministral-3-3B-Instruct & Bactrian-X & \textbf{.336} & \textbf{.364} & .392 & .253 & \textbf{.335} & \textbf{.386} & \textbf{.377} & \textbf{.336} & \textbf{.396} & .271 & \textbf{.254} & \textbf{.354} & \textbf{.392} & \textbf{.301} & \textbf{.410} & \textbf{.311} & \textbf{.409} & \textbf{.303} & \textbf{.370} & \textbf{.351} \\
 & \datasettablee{} & .283 & .322 & \textbf{.427} & \textbf{.255} & .329 & .348 & .336 & .324 & .386 & \textbf{.284} & .239 & .328 & .320 & .275 & .364 & .301 & .362 & .275 & .314 & .305 \\
\midrule
Llama-3.2-3B-Instruct & Bactrian-X & .340 & \textbf{.451} & .423 & .200 & \textbf{.322} & .433 & \textbf{.438} & .340 & \textbf{.457} & \textbf{.223} & .198 & \textbf{.343} & .425 & \textbf{.320} & \textbf{.460} & \textbf{.331} & \textbf{.366} & \textbf{.303} & \textbf{.405} & .248 \\
 & \datasettablee{} & \textbf{.349} & .427 & \textbf{.445} & \textbf{.226} & .311 & \textbf{.440 }& .399 & \textbf{.345} & .455 & .210 & \textbf{.218} & .330 & \textbf{.434} & .316 & .451 & .323 & .351 & .269 & .375 & \textbf{.283} \\
\bottomrule
\end{tabular}%
}
\nopagebreak[4]
\captionof{table}{Pairwise comparison of Bactrian-X and \dataset{} after language-specific fine-tuning on \benchmark{} accuracy over their shared-language subset.}
\label{tab:bactrianx-pairwise}

\clearpage
\subsection{Pairwise Comparisons with Existing Alpaca-Derived Resources on Aya Evaluation Suite with ROUGE-L}
\label{app:external-pairwise-aya-results}
\label{app:external-pairwise-aya-rouge-results}

\subsubsection{Okapi}
\nopagebreak[4]
\vspace{1.35\baselineskip}
\centering
\tiny
\setlength{\tabcolsep}{1.35pt}
\renewcommand{\arraystretch}{0.90}
\resizebox{0.93\textwidth}{!}{%
\begin{tabular}{llrrrrrrrrrrrrrrrr}
\toprule
\textbf{Model} & \textbf{Training resource} & \textbf{ca} & \textbf{da} & \textbf{de} & \textbf{es} & \textbf{eu} & \textbf{fr} & \textbf{hu} & \textbf{it} & \textbf{nl} & \textbf{pt} & \textbf{ro} & \textbf{ru} & \textbf{sk} & \textbf{sr} & \textbf{sv} & \textbf{uk} \\
\midrule
Tiny-Aya-Global & Okapi & .276 & \textbf{.288} & \textbf{.268} & .311 & .179 & \textbf{.289} & .282 & \textbf{.305} & .280 & \textbf{.319} & \textbf{.279} & .110 & .252 & .102 & \textbf{.284} & .137 \\
 & \datasettablee{} & \textbf{.285} & .280 & .258 & \textbf{.312} & \textbf{.188} & .285 & \textbf{.287} & .288 & \textbf{.286} & .311 & .273 & \textbf{.147} & \textbf{.264} & \textbf{.135} & .281 & \textbf{.148} \\
\midrule
Phi-4-Mini-Instruct & Okapi & .266 & \textbf{.275} & .259 & \textbf{.313} & \textbf{.158} & \textbf{.290} & \textbf{.262} & \textbf{.283} & \textbf{.287} & \textbf{.312} & .238 & .107 & .211 & .116 & \textbf{.281} & .119 \\
 & \datasettablee{} & \textbf{.276} & .267 & \textbf{.267} & .300 & .148 & .282 & .261 & .267 & .276 & .310 & \textbf{.250} & \textbf{.128} & \textbf{.216} & \textbf{.144} & .271 & \textbf{.141} \\
\midrule
Ministral-3-3B-Instruct & Okapi & .266 & .249 & .231 & .283 & .142 & .272 & \textbf{.242} & .271 & .256 & .286 & .230 & .093 & .212 & .087 & .253 & .106 \\
 & \datasettablee{} & \textbf{.272} & \textbf{.272} & \textbf{.246} & \textbf{.287} & \textbf{.154} & \textbf{.286} & .239 & \textbf{.273} & \textbf{.263} & \textbf{.298} & \textbf{.257} & \textbf{.113} & \textbf{.227} & \textbf{.117} & \textbf{.276} & \textbf{.132} \\
\midrule
Llama-3.2-3B-Instruct & Okapi & .273 & .262 & .251 & .285 & \textbf{.165} & \textbf{.280} & .252 & .285 & .275 & .281 & .256 & \textbf{.099} & .201 & .074 & .263 & .121 \\
 & \datasettablee{} & \textbf{.283} & \textbf{.271} & \textbf{.254} & \textbf{.302} & .164 & .267 & \textbf{.264} & \textbf{.288} & \textbf{.281} & \textbf{.303} & \textbf{.264} & .086 & \textbf{.210} & \textbf{.121} & \textbf{.275} & \textbf{.127} \\
\bottomrule
\end{tabular}%
}
\nopagebreak[4]
\captionof{table}{Pairwise comparison of Okapi and \dataset{} after language-specific fine-tuning on Aya Evaluation Suite with ROUGE-L over their shared-language subset.}
\label{tab:aya-okapi-rouge-all}
\vspace{-0.4em}

\par\medskip
\subsubsection{MITS}
\nopagebreak[4]
\vspace{1.35\baselineskip}
\centering
\tiny
\setlength{\tabcolsep}{1.35pt}
\renewcommand{\arraystretch}{0.90}
\resizebox{0.98\textwidth}{!}{%
\begin{tabular}{llrrrrrrrrrrrrrrrrrr}
\toprule
\textbf{Model} & \textbf{Training resource} & \textbf{bg} & \textbf{ca} & \textbf{cs} & \textbf{cy} & \textbf{da} & \textbf{de} & \textbf{el} & \textbf{es} & \textbf{et} & \textbf{eu} & \textbf{fi} & \textbf{fr} & \textbf{ga} & \textbf{gd} & \textbf{gl} & \textbf{hu} & \textbf{is} & \textbf{it} \\
\midrule
Tiny-Aya-Global & MITS & \textbf{.123} & .267 & .246 & .240 & .274 & .255 & \textbf{.160} & .295 & .220 & .174 & .207 & .274 & .237 & .192 & .287 & .273 & .207 & .280 \\
 & \datasettablee{} & .122 & \textbf{.285} & \textbf{.255} & \textbf{.259} & \textbf{.280} & \textbf{.258} & \textbf{.160} & \textbf{.312} & \textbf{.232} & \textbf{.188} & \textbf{.212} & \textbf{.285} & \textbf{.251} & \textbf{.216} & \textbf{.294} & \textbf{.287} & \textbf{.220} & \textbf{.288} \\
\midrule
Phi-4-Mini-Instruct & MITS & \textbf{.125} & .259 & .225 & .182 & .264 & .249 & .145 & \textbf{.304} & .153 & .138 & .187 & .274 & .194 & .178 & .284 & .236 & .181 & .252 \\
 & \datasettablee{} & .117 & \textbf{.276} & \textbf{.241} & \textbf{.200} & \textbf{.267} & \textbf{.267} & \textbf{.157} & .300 & \textbf{.172} & \textbf{.148} & \textbf{.200} & \textbf{.282} & \textbf{.195} & \textbf{.203} & \textbf{.293} & \textbf{.261} & \textbf{.200} & \textbf{.267} \\
\midrule
Ministral-3-3B-Instruct & MITS & \textbf{.115} & .261 & .232 & .196 & .243 & .233 & .138 & .285 & .191 & .147 & .187 & .266 & .192 & .185 & .265 & .223 & .214 & .253 \\
 & \datasettablee{} & .100 & \textbf{.272} & \textbf{.254} & \textbf{.214} & \textbf{.272} & \textbf{.246} & \textbf{.153} & \textbf{.287} & \textbf{.199} & \textbf{.154} & \textbf{.203} & \textbf{.286} & \textbf{.197} & \textbf{.186} & \textbf{.281} & \textbf{.239} & \textbf{.221} & \textbf{.273} \\
\midrule
Llama-3.2-3B-Instruct & MITS & \textbf{.115} & .270 & \textbf{.245} & .196 & .267 & .234 & \textbf{.139} & .291 & .183 & .145 & \textbf{.184} & \textbf{.282} & \textbf{.204} & .198 & .276 & .251 & .196 & .282 \\
 & \datasettablee{} & .112 & \textbf{.283} & .241 & \textbf{.203} & \textbf{.271} & \textbf{.254} & .133 & \textbf{.302} & \textbf{.193} & \textbf{.164} & .183 & .267 & .188 & \textbf{.215} & \textbf{.291} & \textbf{.264} & \textbf{.202} & \textbf{.288} \\
\bottomrule
\end{tabular}%
}
\nopagebreak[4]
\captionof{table}{Pairwise comparison of MITS and \dataset{} after language-specific fine-tuning on Aya Evaluation Suite with ROUGE-L over their shared-language subset (languages 1--18 of 35).}
\label{tab:aya-mits-rouge-a}

\vspace{1.35\baselineskip}
\centering
\tiny
\setlength{\tabcolsep}{1.35pt}
\renewcommand{\arraystretch}{0.90}
\resizebox{0.98\textwidth}{!}{%
\begin{tabular}{llrrrrrrrrrrrrrrrrr}
\toprule
\textbf{Model} & \textbf{Training resource} & \textbf{lb} & \textbf{lt} & \textbf{lv} & \textbf{mk} & \textbf{mt} & \textbf{nl} & \textbf{nob} & \textbf{pl} & \textbf{pt} & \textbf{ro} & \textbf{ru} & \textbf{sk} & \textbf{sl} & \textbf{sq} & \textbf{sr} & \textbf{sv} & \textbf{uk} \\
\midrule
Tiny-Aya-Global & MITS & .204 & .203 & .211 & \textbf{.142} & .142 & .268 & .268 & .229 & .289 & .256 & .141 & .248 & .239 & .224 & .120 & .270 & \textbf{.157} \\
 & \datasettablee{} & \textbf{.221} & \textbf{.213} & \textbf{.223} & .134 & \textbf{.260} & \textbf{.286} & \textbf{.291} & \textbf{.238} & \textbf{.311} & \textbf{.273} & \textbf{.147} & \textbf{.264} & \textbf{.255} & \textbf{.251} & \textbf{.135} & \textbf{.281} & .148 \\
\midrule
Phi-4-Mini-Instruct & MITS & .214 & .172 & .183 & .132 & \textbf{.176} & .259 & .271 & .217 & .298 & .233 & .127 & .208 & .210 & .223 & .122 & .261 & \textbf{.152} \\
 & \datasettablee{} & \textbf{.227} & \textbf{.195} & \textbf{.184} & \textbf{.137} & .169 & \textbf{.276} & \textbf{.283} & \textbf{.227} & \textbf{.310} & \textbf{.250} & \textbf{.128} & \textbf{.216} & \textbf{.211} & \textbf{.233} & \textbf{.144} & \textbf{.271} & .141 \\
\midrule
Ministral-3-3B-Instruct & MITS & .207 & .183 & .183 & \textbf{.128} & .183 & .245 & .251 & .209 & .278 & .242 & \textbf{.121} & .203 & .194 & .231 & .113 & .259 & \textbf{.142} \\
 & \datasettablee{} & \textbf{.209} & \textbf{.191} & \textbf{.195} & .126 & \textbf{.202} & \textbf{.263} & \textbf{.263} & \textbf{.224} & \textbf{.298} & \textbf{.257} & .113 & \textbf{.227} & \textbf{.212} & \textbf{.239} & \textbf{.117} & \textbf{.276} & .132 \\
\midrule
Llama-3.2-3B-Instruct & MITS & .222 & \textbf{.203} & .181 & \textbf{.129} & .187 & .263 & .259 & .214 & .292 & .240 & \textbf{.106} & .205 & .209 & .217 & .114 & .265 & \textbf{.131} \\
 & \datasettablee{} & \textbf{.227} & .202 & \textbf{.182} & .120 & \textbf{.199} & \textbf{.281} & \textbf{.267} & \textbf{.222} & \textbf{.303} & \textbf{.264} & .086 & \textbf{.210} & \textbf{.210} & \textbf{.257} & \textbf{.121} & \textbf{.275} & .127 \\
\bottomrule
\end{tabular}%
}
\nopagebreak[4]
\captionof{table}{Pairwise comparison of MITS and \dataset{} after language-specific fine-tuning on Aya Evaluation Suite with ROUGE-L over their shared-language subset (languages 19--35 of 35).}
\label{tab:aya-mits-rouge-b}
\vspace{-0.4em}
\par\medskip
\subsubsection{Bactrian-X}
\nopagebreak[4]
\vspace{1\baselineskip}
\centering
\tiny
\setlength{\tabcolsep}{1.35pt}
\renewcommand{\arraystretch}{0.90}
\resizebox{0.98\textwidth}{!}{%
\begin{tabular}{llrrrrrrrrrrrrrrrrrrr}
\toprule
\textbf{Model} & \textbf{Training resource} & \textbf{cs} & \textbf{de} & \textbf{es} & \textbf{et} & \textbf{fi} & \textbf{fr} & \textbf{gl} & \textbf{it} & \textbf{lt} & \textbf{lv} & \textbf{mk} & \textbf{nl} & \textbf{pl} & \textbf{pt} & \textbf{ro} & \textbf{ru} & \textbf{sl} & \textbf{sv} & \textbf{uk} \\
\midrule
Tiny-Aya-Global & Bactrian-X & .127 & .113 & .304 & .218 & \textbf{.213} & .283 & .273 & \textbf{.292} & \textbf{.222} & .218 & .116 & .275 & .228 & .299 & .122 & .117 & .242 & \textbf{.290} & \textbf{.158} \\
 & \datasettablee{} & \textbf{.255} & \textbf{.258} & \textbf{.312} & \textbf{.232} & .212 & \textbf{.285} & \textbf{.294} & .288 & .213 & \textbf{.223} & \textbf{.134} & \textbf{.286} & \textbf{.238} & \textbf{.311} & \textbf{.273} & \textbf{.147} & \textbf{.255} & .281 & .148 \\
\midrule
Phi-4-Mini-Instruct & Bactrian-X & .236 & .250 & \textbf{.303} & \textbf{.176} & .198 & .281 & .248 & .263 & .184 & .180 & .116 & .271 & .217 & .306 & .243 & .118 & .196 & \textbf{.276} & \textbf{.147} \\
 & \datasettablee{} & \textbf{.241} & \textbf{.267} & .300 & .172 & \textbf{.200} & \textbf{.282} & \textbf{.293} & \textbf{.267} & \textbf{.195} & \textbf{.184} & \textbf{.137} & \textbf{.276} & \textbf{.227} & \textbf{.310} & \textbf{.250} & \textbf{.128} & \textbf{.211} & .271 & .141 \\
\midrule
Ministral-3-3B-Instruct & Bactrian-X & .239 & .236 & \textbf{.289} & .179 & .190 & .262 & .238 & .265 & .178 & .177 & .124 & .254 & .218 & .276 & .243 & .106 & .196 & .258 & .130 \\
 & \datasettablee{} & \textbf{.254} & \textbf{.246} & .287 & \textbf{.199} & \textbf{.203} & \textbf{.286} & \textbf{.281} & \textbf{.273} & \textbf{.191} & \textbf{.195} & \textbf{.126} & \textbf{.263} & \textbf{.224} & \textbf{.298} & \textbf{.257} & \textbf{.113} & \textbf{.212} & \textbf{.276} & \textbf{.132} \\
\midrule
Llama-3.2-3B-Instruct & Bactrian-X & .224 & .241 & .295 & .191 & .173 & .257 & .255 & .273 & .180 & .180 & .109 & .276 & .211 & .289 & .255 & \textbf{.111} & .200 & .258 & \textbf{.127} \\
 & \datasettablee{} & \textbf{.241} & \textbf{.254} & \textbf{.302} & \textbf{.193} & \textbf{.183} & \textbf{.267} & \textbf{.291} & \textbf{.288} & \textbf{.202} & \textbf{.182} & \textbf{.120} & \textbf{.281} & \textbf{.222} & \textbf{.303} & \textbf{.264} & .086 & \textbf{.210} & \textbf{.275} & \textbf{.127} \\
\bottomrule
\end{tabular}%
}
\nopagebreak[4]
\captionof{table}{Pairwise comparison of Bactrian-X and \dataset{} after language-specific fine-tuning on Aya Evaluation Suite with ROUGE-L over their shared-language subset.}
\label{tab:aya-bactrianx-rouge-all}

\clearpage
\subsection{Pairwise Comparisons with Existing Alpaca-Derived Resources on Aya Evaluation Suite with F\textsc{bert}}
\label{app:external-pairwise-aya-bertscore-results}

\subsubsection{Okapi}
\nopagebreak[4]
\vspace{1.35\baselineskip}
\centering
\tiny
\setlength{\tabcolsep}{1.35pt}
\renewcommand{\arraystretch}{0.90}
\resizebox{0.93\textwidth}{!}{%
\begin{tabular}{llrrrrrrrrrrrrrrrr}
\toprule
\textbf{Model} & \textbf{Training resource} & \textbf{ca} & \textbf{da} & \textbf{de} & \textbf{es} & \textbf{eu} & \textbf{fr} & \textbf{hu} & \textbf{it} & \textbf{nl} & \textbf{pt} & \textbf{ro} & \textbf{ru} & \textbf{sk} & \textbf{sr} & \textbf{sv} & \textbf{uk} \\
\midrule
Tiny-Aya-Global & Okapi & .881 & \textbf{.892} & \textbf{.888} & \textbf{.891} & .866 & \textbf{.888} & \textbf{.885} & \textbf{.893} & \textbf{.887} & \textbf{.892} & \textbf{.886} & .888 & \textbf{.885} & .875 & \textbf{.888} & \textbf{.884} \\
 & \datasettablee{} & \textbf{.883} & .887 & .886 & .890 & \textbf{.869} & .887 & \textbf{.885} & .887 & \textbf{.887} & .889 & .882 & \textbf{.889} & \textbf{.885} & \textbf{.881} & .887 & \textbf{.884} \\
\midrule
Phi-4-Mini-Instruct & Okapi & .877 & \textbf{.886} & \textbf{.887} & \textbf{.892} & \textbf{.844} & \textbf{.888} & \textbf{.877} & \textbf{.886} & \textbf{.887} & \textbf{.891} & \textbf{.874} & \textbf{.888} & \textbf{.870} & .865 & \textbf{.885} & \textbf{.878} \\
 & \datasettablee{} & \textbf{.878} & .884 & .885 & .888 & .839 & .885 & .872 & .884 & .883 & .888 & .872 & .887 & .867 & \textbf{.873} & .882 & \textbf{.878} \\
\midrule
Ministral-3-3B-Instruct & Okapi & \textbf{.877} & .877 & .877 & \textbf{.885} & .843 & .882 & \textbf{.864} & \textbf{.884} & .878 & .883 & .872 & .881 & .866 & .866 & .877 & .871 \\
 & \datasettablee{} & .876 & \textbf{.882} & \textbf{.882} & \textbf{.885} & \textbf{.849} & \textbf{.886} & .861 & .883 & \textbf{.879} & \textbf{.884} & \textbf{.873} & \textbf{.884} & \textbf{.868} & \textbf{.872} & \textbf{.882} & \textbf{.875} \\
\midrule
Llama-3.2-3B-Instruct & Okapi & .878 & \textbf{.881} & \textbf{.884} & .888 & .847 & \textbf{.885} & .871 & .887 & \textbf{.885} & .886 & .875 & \textbf{.883} & .862 & .856 & \textbf{.883} & .875 \\
 & \datasettablee{} & \textbf{.881} & .880 & \textbf{.884} & \textbf{.890} & \textbf{.851} & .882 & \textbf{.876} & \textbf{.888} & .884 & \textbf{.889} & \textbf{.881} & .881 & \textbf{.866} & \textbf{.869} & .879 & \textbf{.878} \\
\bottomrule
\end{tabular}%
}
\nopagebreak[4]
\captionof{table}{Pairwise comparison of Okapi and \dataset{} after language-specific fine-tuning on Aya Evaluation Suite with F\textsc{bert} over their shared-language subset.}
\label{tab:aya-okapi-bertscore-all}
\vspace{-0.4em}

\par\medskip
\subsubsection{MITS}
\nopagebreak[4]
\vspace{1.35\baselineskip}
\centering
\tiny
\setlength{\tabcolsep}{1.35pt}
\renewcommand{\arraystretch}{0.90}
\resizebox{0.98\textwidth}{!}{%
\begin{tabular}{llrrrrrrrrrrrrrrrrrr}
\toprule
\textbf{Model} & \textbf{Training resource} & \textbf{bg} & \textbf{ca} & \textbf{cs} & \textbf{cy} & \textbf{da} & \textbf{de} & \textbf{el} & \textbf{es} & \textbf{et} & \textbf{eu} & \textbf{fi} & \textbf{fr} & \textbf{ga} & \textbf{gd} & \textbf{gl} & \textbf{hu} & \textbf{is} & \textbf{it} \\
\midrule
Tiny-Aya-Global & MITS & .879 & .878 & .877 & .862 & .883 & .883 & .879 & .886 & .869 & .862 & .873 & .883 & .853 & .824 & .880 & .879 & .847 & .885 \\
 & \datasettablee{} & \textbf{.886} & \textbf{.883} & \textbf{.882} & \textbf{.867} & \textbf{.887} & \textbf{.886} & \textbf{.887} & \textbf{.890} & \textbf{.877} & \textbf{.869} & \textbf{.878} & \textbf{.887} & \textbf{.857} & \textbf{.835} & \textbf{.885} & \textbf{.885} & \textbf{.855} & \textbf{.887} \\
\midrule
Phi-4-Mini-Instruct & MITS & .869 & .870 & .867 & .836 & .880 & .881 & .863 & .887 & .836 & .832 & .863 & .882 & .830 & .820 & .877 & .864 & .836 & .878 \\
 & \datasettablee{} & \textbf{.876} & \textbf{.878} & \textbf{.874} & \textbf{.840} & \textbf{.884} & \textbf{.885} & \textbf{.867} & \textbf{.888} & \textbf{.850} & \textbf{.839} & \textbf{.872} & \textbf{.885} & \textbf{.832} & \textbf{.827} & \textbf{.879} & \textbf{.872} & \textbf{.845} & \textbf{.884} \\
\midrule
Ministral-3-3B-Instruct & MITS & .865 & .871 & .869 & .836 & .874 & .874 & .871 & .882 & .852 & .838 & .862 & .880 & .822 & .815 & .872 & .855 & .839 & .876 \\
 & \datasettablee{} & \textbf{.878} & \textbf{.876} & \textbf{.876} & \textbf{.842} & \textbf{.882} & \textbf{.882} & \textbf{.879} & \textbf{.885} & \textbf{.859} & \textbf{.849} & \textbf{.872} & \textbf{.886} & \textbf{.831} & \textbf{.822} & \textbf{.878} & \textbf{.861} & \textbf{.852} & \textbf{.883} \\
\midrule
Llama-3.2-3B-Instruct & MITS & .870 & .875 & .869 & .843 & .878 & .877 & .867 & .886 & .855 & .837 & .858 & \textbf{.883} & .832 & .831 & .876 & .868 & .840 & .882 \\
 & \datasettablee{} & \textbf{.875} & \textbf{.881} & \textbf{.873} & \textbf{.848} & \textbf{.880} & \textbf{.884} & \textbf{.877} & \textbf{.890} & \textbf{.861} & \textbf{.851} & \textbf{.869} & .882 & \textbf{.833} & \textbf{.839} & \textbf{.882} & \textbf{.876} & \textbf{.849} & \textbf{.888} \\
\bottomrule
\end{tabular}%
}
\nopagebreak[4]
\captionof{table}{Pairwise comparison of MITS and \dataset{} after language-specific fine-tuning on Aya Evaluation Suite with F\textsc{bert} over their shared-language subset (languages 1--18 of 35).}
\label{tab:aya-mits-bertscore-a}

\vspace{1.35\baselineskip}
\centering
\tiny
\setlength{\tabcolsep}{1.35pt}
\renewcommand{\arraystretch}{0.90}
\resizebox{0.98\textwidth}{!}{%
\begin{tabular}{llrrrrrrrrrrrrrrrrr}
\toprule
\textbf{Model} & \textbf{Training resource} & \textbf{lb} & \textbf{lt} & \textbf{lv} & \textbf{mk} & \textbf{mt} & \textbf{nl} & \textbf{nob} & \textbf{pl} & \textbf{pt} & \textbf{ro} & \textbf{ru} & \textbf{sk} & \textbf{sl} & \textbf{sq} & \textbf{sr} & \textbf{sv} & \textbf{uk} \\
\midrule
Tiny-Aya-Global & MITS & .825 & .869 & .870 & .873 & .854 & .882 & .885 & .876 & .886 & .877 & .883 & .878 & .877 & .858 & .874 & .882 & .878 \\
 & \datasettablee{} & \textbf{.835} & \textbf{.875} & \textbf{.874} & \textbf{.878} & \textbf{.858} & \textbf{.887} & \textbf{.890} & \textbf{.882} & \textbf{.889} & \textbf{.882} & \textbf{.889} & \textbf{.885} & \textbf{.884} & \textbf{.868} & \textbf{.881} & \textbf{.887} & \textbf{.884} \\
\midrule
Phi-4-Mini-Instruct & MITS & .825 & .848 & .849 & .864 & .828 & .878 & .879 & .869 & .884 & .864 & .882 & .861 & .861 & .853 & .866 & .876 & .873 \\
 & \datasettablee{} & \textbf{.832} & \textbf{.857} & \textbf{.853} & \textbf{.872} & \textbf{.830} & \textbf{.883} & \textbf{.886} & \textbf{.876} & \textbf{.888} & \textbf{.872} & \textbf{.887} & \textbf{.867} & \textbf{.865} & \textbf{.858} & \textbf{.873} & \textbf{.882} & \textbf{.878} \\
\midrule
Ministral-3-3B-Instruct & MITS & .825 & .857 & .849 & .862 & .830 & .872 & .874 & .869 & .878 & .868 & .876 & .861 & .859 & .854 & .868 & .873 & .868 \\
 & \datasettablee{} & \textbf{.826} & \textbf{.862} & \textbf{.856} & \textbf{.872} & \textbf{.839} & \textbf{.879} & \textbf{.880} & \textbf{.875} & \textbf{.884} & \textbf{.873} & \textbf{.884} & \textbf{.868} & \textbf{.870} & \textbf{.860} & \textbf{.872} & \textbf{.882} & \textbf{.875} \\
\midrule
Llama-3.2-3B-Instruct & MITS & .831 & .857 & .853 & .861 & .834 & .878 & .877 & .865 & .887 & .869 & .878 & .860 & .863 & .851 & .864 & .877 & .872 \\
 & \datasettablee{} & \textbf{.837} & \textbf{.867} & \textbf{.858} & \textbf{.872} & \textbf{.840} & \textbf{.884} & \textbf{.880} & \textbf{.877} & \textbf{.889} & \textbf{.881} & \textbf{.881} & \textbf{.866} & \textbf{.869} & \textbf{.867} & \textbf{.869} & \textbf{.879} & \textbf{.878} \\
\bottomrule
\end{tabular}%
}
\nopagebreak[4]
\captionof{table}{Pairwise comparison of MITS and \dataset{} after language-specific fine-tuning on Aya Evaluation Suite with F\textsc{bert} over their shared-language subset (languages 19--35 of 35).}
\label{tab:aya-mits-bertscore-b}
\vspace{-0.4em}
\par\medskip
\subsubsection{Bactrian-X}
\nopagebreak[4]
\vspace{1.35\baselineskip}
\centering
\tiny
\setlength{\tabcolsep}{1.35pt}
\renewcommand{\arraystretch}{0.90}
\resizebox{0.98\textwidth}{!}{%
\begin{tabular}{llrrrrrrrrrrrrrrrrrrr}
\toprule
\textbf{Model} & \textbf{Training resource} & \textbf{cs} & \textbf{de} & \textbf{es} & \textbf{et} & \textbf{fi} & \textbf{fr} & \textbf{gl} & \textbf{it} & \textbf{lt} & \textbf{lv} & \textbf{mk} & \textbf{nl} & \textbf{pl} & \textbf{pt} & \textbf{ro} & \textbf{ru} & \textbf{sl} & \textbf{sv} & \textbf{uk} \\
\midrule
Tiny-Aya-Global & Bactrian-X & .838 & .831 & .888 & .871 & \textbf{.878} & .885 & .880 & .886 & .871 & .871 & .870 & .883 & .876 & .887 & .832 & .886 & .877 & .885 & .880 \\
 & \datasettablee{} & \textbf{.882} & \textbf{.886} & \textbf{.890} & \textbf{.877} & \textbf{.878} & \textbf{.887} & \textbf{.885} & \textbf{.887} & \textbf{.875} & \textbf{.874} & \textbf{.878} & \textbf{.887} & \textbf{.882} & \textbf{.889} & \textbf{.882} & \textbf{.889} & \textbf{.884} & \textbf{.887} & \textbf{.884} \\
\midrule
Phi-4-Mini-Instruct & Bactrian-X & \textbf{.875} & .883 & \textbf{.888} & .849 & .870 & .884 & .872 & .878 & .854 & .850 & .867 & .880 & .872 & .887 & .869 & .886 & .855 & .879 & .876 \\
 & \datasettablee{} & .874 & \textbf{.885} & \textbf{.888} & \textbf{.850} & \textbf{.872} & \textbf{.885} & \textbf{.879} & \textbf{.884} & \textbf{.857} & \textbf{.853} & \textbf{.872} & \textbf{.883} & \textbf{.876} & \textbf{.888} & \textbf{.872} & \textbf{.887} & \textbf{.865} & \textbf{.882} & \textbf{.878} \\
\midrule
Ministral-3-3B-Instruct & Bactrian-X & .872 & .878 & \textbf{.885} & .849 & .866 & .879 & .869 & .880 & .855 & .851 & .865 & .876 & .872 & .881 & .870 & .876 & .856 & .870 & .869 \\
 & \datasettablee{} & \textbf{.876} & \textbf{.882} & .885 & \textbf{.859} & \textbf{.872} & \textbf{.886} & \textbf{.878} & \textbf{.883} & \textbf{.862} & \textbf{.856} & \textbf{.872} & \textbf{.879} & \textbf{.875} & \textbf{.884} & \textbf{.873} & \textbf{.884} & \textbf{.870} & \textbf{.882} & \textbf{.875} \\
\midrule
Llama-3.2-3B-Instruct & Bactrian-X & .867 & .881 & .887 & .854 & .861 & .880 & .876 & .881 & .852 & .851 & .866 & .880 & .869 & .887 & .874 & .877 & .862 & \textbf{.880} & .868 \\
 & \datasettablee{} & \textbf{.873} & \textbf{.884} & \textbf{.890} & \textbf{.861} & \textbf{.869} & \textbf{.882} & \textbf{.882} & \textbf{.888} & \textbf{.867} & \textbf{.858} & \textbf{.872} & \textbf{.884} & \textbf{.877} & \textbf{.889} & \textbf{.881} & \textbf{.881} & \textbf{.869} & .879 & \textbf{.878} \\
\bottomrule
\end{tabular}%
}
\nopagebreak[4]
\captionof{table}{Pairwise comparison of Bactrian-X and \dataset{} after language-specific fine-tuning on Aya Evaluation Suite with F\textsc{bert} over their shared-language subset.}
\label{tab:aya-bactrianx-bertscore-all}

\FloatBarrier
\clearpage
\twocolumn
\clearpage
\FloatBarrier
\clearpage
\twocolumn

\normalsize
\justifying

\section{Reproducibility Details}
\label{app:reproducibility}

\subsection{Checkpoint Identifiers}
\label{app:checkpoints}

The four evaluated checkpoints are:
\vspace{-0.17cm}
\begin{itemize}
    \setlength{\itemsep}{1pt}
    \setlength{\parsep}{0pt}
    \setlength{\parskip}{0pt}
    \setlength{\topsep}{3pt}
    \setlength{\partopsep}{0pt}
    \small
    \item \nolinkurl{CohereLabs/tiny-aya-global}
    \item \nolinkurl{microsoft/Phi-4-mini-instruct}
    \item \nolinkurl{mistralai/Ministral-3-3B-Instruct-2512}
    \item \nolinkurl{meta-llama/Llama-3.2-3B-Instruct}
\end{itemize}

\subsection{Fine-Tuning Hyperparameters}
\label{app:hyperparameters}

All language-specific runs use the same optimisation and adapter
configuration, reported in Table~\ref{tab:hyperparameters}.

\begin{table}[H]
\centering
\small
\setlength{\tabcolsep}{3.5pt}
\renewcommand{\arraystretch}{1.02}

\begin{tabularx}{\columnwidth}{
    @{}
    >{\raggedright\arraybackslash}p{0.43\columnwidth}
    >{\raggedright\arraybackslash}X
    @{}
}
\toprule
\textbf{Parameter} & \textbf{Value} \\
\midrule
Method & LoRA \\
Rank / alpha / dropout & 16 / 32 / 0.05 \\
Target modules & \texttt{q}, \texttt{k}, \texttt{v}, and \texttt{o} projections \\
Maximum sequence length & 1,300 tokens by default \\
Optimiser & AdamW \\
Learning rate & $1 \times 10^{-5}$ \\
Micro-batch size & 2 \\
Gradient-accumulation steps & 32 \\
Effective batch size & 64 \\
Gradient clipping & 1.0 \\
Precision & bf16 \\
Random seed & 42 \\
Checkpoint selection & Validation loss with early stopping \\
\bottomrule
\end{tabularx}
\caption{Fine-tuning configuration shared by all language-specific runs. Based on token-length analysis, we use a 1,600-token limit for \texttt{ga}, \texttt{gd}, \texttt{is}, \texttt{mk}, \texttt{mt} and \texttt{sq} to reduce truncation in the upper tail of the sequence-length
distribution.}
\label{tab:hyperparameters}
\end{table}

\subsubsection{Prompt Formatting and Language-Specific Alpaca Templates}
\label{app:finetuning-prompts}

For fine-tuning, each example was serialised using the model-specific chat template provided by each model’s tokeniser, ensuring compatibility with the format expected by each model. The user turn was constructed with an Alpaca-style prompt in the same language as the instruction.

We translated the original English Alpaca template into each target language using ChatGPT. In every version, we modified the original instruction, \textit{Write a response that appropriately completes the request}, by adding an explicit response-language requirement. For English, this yielded \textit{Write a response in English that appropriately completes the request}. Equivalent specifications, such as \textit{auf Deutsch}, \textit{en català} and \textit{in italiano}, were added for the other languages. This language requirement is not present in the original Alpaca template. Each example was therefore paired with a template written in the same language as the instruction and explicitly requesting a response in that language.

Following the original format, we used two template variants. Records with a non-empty input included separate \texttt{\midsize Instruction}, \texttt{\midsize Input} and \texttt{\midsize Response} blocks, whereas records without an input omitted the \texttt{\midsize Input} block.

\vspace{0.5em}

\noindent\textbf{English template}\par
\nobreak
\begin{lstlisting}[
  style=prompt,
  breaklines=true,
  aboveskip=0.3em,
  belowskip=0.8em
]
Below is an instruction that describes a task, paired with an input that provides further context. Write a response in English that appropriately completes the request.

### Instruction:
{instruction text}

### Input:
{optional input text}

### Response:
{expected output}
\end{lstlisting}

\vspace{0.25em}
\noindent
Examples in other languages:\par
\smallskip

\noindent\textbf{German template}\par
\nobreak
\begin{lstlisting}[
  style=prompt,
  breaklines=true,
  aboveskip=0.3em,
  belowskip=0.8em,
  literate=
    {ä}{{\"a}}1
    {ö}{{\"o}}1
    {ü}{{\"u}}1
    {Ä}{{\"A}}1
    {Ö}{{\"O}}1
    {Ü}{{\"U}}1
    {ß}{{\ss}}1
]
Nachfolgend finden Sie eine Anweisung, die eine Aufgabe beschreibt. Schreiben Sie eine Antwort auf Deutsch, die die Anfrage angemessen erfüllt.

### Anweisung:
{instruction text}

### Eingabe:
{optional input text}

### Antwort:
{expected output}
\end{lstlisting}

\noindent\textbf{Catalan template}\par
\nobreak
\begin{lstlisting}[
  style=prompt,
  breaklines=true,
  aboveskip=0.3em,
  belowskip=0.8em,
  literate=
    {à}{{\`a}}1
    {é}{{\'e}}1
    {ó}{{\'o}}1
    {ç}{{\c c}}1
    {·}{{\textperiodcentered}}1
]
A continuació es mostra una instrucció que descriu una tasca, juntament amb una entrada que proporciona més context. Escriviu una resposta en català que completi adequadament la sol·licitud.

### Instrucció:
{instruction text}

### Entrada:
{optional input text}

### Resposta:
{expected output}
\end{lstlisting}

\subsection{Gemma-4 Decision-Module Configuration}
\begin{table}[H]
\centering
\small
\begin{tabularx}{\columnwidth}{
    @{}
    >{\raggedright\arraybackslash}p{0.37\columnwidth}
    >{\raggedright\arraybackslash}X
    @{}
}
\toprule
\textbf{Component} & \textbf{Configuration} \\
\midrule
Decision model
    & Gemma-4-31B-IT, \texttt{Q8\_K\_XL} GGUF \\
Draft model
    & Gemma-4-E2B-IT, \texttt{Q4\_K\_XL} GGUF \\
Decoding
    & Temperature $=0$, top-$p=0.95$, min-$p=0.01$ \\
Penalties
    & None explicitly configured \\
Reasoning
    & Disabled \\
Speculative decoding
    & 5--16 draft tokens \\
Prompt formatting
    & Gemma-4 Jinja template \\
Context / parallelism
    & 8,912 tokens / 2 server slots \\
KV cache / attention
    & \texttt{q8\_0}; unified KV cache; Flash Attention \\
Batch / physical batch
    & 1,024 / 512 \\
GPU offloading
    & Up to 99 main- and draft-model layers \\
\bottomrule
\end{tabularx}
\caption{Inference and decoding configuration of the Gemma-4 decision module.}
\label{tab:gemma-inference}
\end{table}
%\label{app:aggregation-rules}

%For \benchmark{}, each model--condition result is computed as the arithmetic mean over the available target-language scores. Overall values first average the languages within each model and then average the resulting four model-level means. This procedure prevents differences in language coverage from implicitly assigning greater weight to one model.

%For the Aya Evaluation Suite, English is excluded from aggregate comparisons because English results are unavailable for the Direct MT and Task-Preserved conditions. Missing metric values are omitted rather than imputed.

\newpage
\subsection{Gemma-4 Prompt Templates}
\label{app:gemma-prompts}

This section reports the system and user prompt templates used by the Gemma-4 decision module for semantic annotation, sample-level routing, field-level policy prediction, task-level localisation and consistency validation. Placeholders enclosed in braces are populated at runtime.

\subsubsection{Semantic Annotation}

\medskip
\noindent\textbf{System prompt}\par
\nobreak
\begin{lstlisting}[style=prompt]
You are a careful dataset annotator. Output JSON only.
\end{lstlisting}

\medskip
\noindent\textbf{User prompt template}\par
\nobreak
\begin{lstlisting}[style=prompt]
You are labelling BOTH the TASK TYPE(S) and the DOMAIN(S) of the request.

You must pick:
- one PRIMARY task type (exactly one)
- zero or more SECONDARY task types (may be empty)
- one PRIMARY domain (exactly one)
- zero or more SECONDARY domains (may be empty)

Allowed task types (use EXACT strings):
question_answering, reasoning, summarisation, translation, paraphrasing, classification, information_extraction, creative_writing, procedural_instructions, coding, style_transfer, evaluation, other

Allowed domain types (use EXACT strings):
general, programming, technology, science, mathematics, medicine, law, business, education, humanities, social_science, creative, other

Rules (tasks):
- primary_task must be exactly one of the allowed task types.
- secondary_tasks must be a JSON array of allowed task types.
- secondary_tasks must NOT repeat primary_task.
- If uncertain, use "other".

Rules (domains):
- primary_domain must be exactly one of the allowed domain types.
- secondary_domains must be a JSON array of allowed domain types.
- secondary_domains must NOT repeat primary_domain.
- Use "general" when the topic is everyday or non-specialised.
- Use "other" only if none fit.

Return ONLY valid JSON with exactly these keys:
{
  "primary_task": "<one_task_type>",
  "secondary_tasks": ["<task_type>", "..."],
  "primary_domain": "<one_domain_type>",
  "secondary_domains": ["<domain_type>", "..."]
}

No extra text. No markdown. No code fences.
---
Below is an instruction that describes a task, paired with an input that provides further context. Write a response that appropriately completes the request.

### Instruction:
{instruction}

### Input:
{input}

### Response:
\end{lstlisting}

\newpage
\subsubsection{Sample-Level MT Suitability Gate}

\medskip
\noindent\textbf{System prompt}\par
\nobreak
\begin{lstlisting}[style=prompt]
You are a strict data curation assistant. Return JSON only.
\end{lstlisting}

\medskip
\noindent\textbf{User prompt template}\par
\nobreak
\begin{lstlisting}[style=prompt]
You are deciding the SAMPLE-LEVEL translation policy for a dataset row.
This first pass is a global gate: decide whether the task can be preserved through field-wise machine translation.
A later field-level pass will decide which individual fields to translate or preserve.
Source language: {source_lang_name}
Target languages: {target_desc_text}
Primary task: {primary_task}
Primary domain: {primary_domain}
Secondary tasks: {secondary_tasks}
Secondary domains: {secondary_domains}

Field values:
instruction: {instruction}
input: {input}
output: {output}

Rules:
- Keep task semantics intact.
- Decide if this sample should be translated to other languages at all.
- Set include_sample=false when the task depends on source-language form rather than meaning.
- Set include_sample=false for language-form tasks (spelling, grammar, tense/voice edits, typo correction).
- Set include_sample=false for token-level/wordplay tasks (letter counts, rhymes/rhyming/rhymes with, homophones, anagrams).
- Set include_sample=false for idioms, proverbs, puns, mnemonics, first-letter tasks, and exact letter/character constraints.
- Set include_sample=false when the output already violates a strict word/line/count constraint.
- For classification, discard figurative-language detection tasks if MT may alter label logic.
- For coding, discard mislabelled language-conjugation/grammar-only tasks.
- For creative writing, discard strict form/sound constraints (haiku/limerick/acrostic/rhyme/wordplay).
- For evaluation, discard English-form correction/judgement tasks (grammar/spelling/punctuation/usage).
- For information extraction, discard grammar/syntax/token-form extraction tasks.
- For other, discard language-form editing/spelling/punctuation/phonetic/accent tasks.
- For paraphrasing, discard grammar/tense/voice/person/word-level constrained rewrites.
- Set include_sample=false for English-constrained generation/rewrite tasks (e.g., 'in formal/simple/modern English', English-sentence-only constraints).
- Set include_sample=false for English alphabet/pangram constraints.
- Set include_sample=false for language-specific tasks (English-only constraints).
- Keep include_sample=true for normal QA, reasoning, coding, summarisation, and creative generation tasks.
- Keep include_sample=true for paraphrasing focused on clarity/conciseness/flow (not language-form mechanics).
- Translation tasks can still be include_sample=true; field policy will protect fixed-language fields.
- If uncertain whether the sample involves a language-form constraint or ordinary meaning-preserving translation, choose include_sample=false.
- If include_sample=true, reason_code must be directly_translatable and evidence must be empty.
- If include_sample=false, choose exactly one reason_code from the allowed list and provide short evidence from the sample.

Allowed reason_code values:
- directly_translatable
- english_grammar_or_spelling
- rhyme_or_sound_constraint
- letter_or_character_constraint
- word_count_or_line_count_constraint
- idiom_or_proverb
- wordplay_pun_anagram_palindrome
- mnemonic_or_initial_letters
- language_specific_task
- output_already_violates_instruction
- other_not_directly_translatable

Return ONLY valid JSON with exactly this schema:
{
  "sample": {
    "include_sample": true,
    "reason_code": "directly_translatable",
    "evidence": ""
  }
}
No markdown. No extra keys.
\end{lstlisting}

\subsubsection{Field-Level Translation Policy}

\medskip
\noindent\textbf{System prompt}\par
\nobreak
\begin{lstlisting}[style=prompt]
You are a strict data curation assistant. Return JSON only.
\end{lstlisting}

\medskip
\noindent\textbf{User prompt template}\par
\nobreak
\begin{lstlisting}[style=prompt]
You are deciding the FIELD-LEVEL translation policy for a dataset row.
This sample has passed the global gate and can be preserved through field-wise machine translation.
This second pass decides which individual fields should be translated or preserved.
Make decisions in one pass for ALL provided target languages, using a default plus language-specific exceptions.
Source language: {source_lang_name}
Target languages: {target_desc_text}
Primary task: {primary_task}
Primary domain: {primary_domain}
Secondary tasks: {secondary_tasks}
Secondary domains: {secondary_domains}

Field values:
instruction: {instruction}
input: {input}
output: {output}

Rules:
- Keep task semantics intact.
- Set should_translate=false only when translation would break
task intent.
- If a field is empty or missing, should_translate=false.
- Do not set false just because task is rewriting/paraphrasing.
- For language-specific tasks, sample-level policy already
filtered unsuitable samples.
- For classification: preserve input for language-ID/code/
formula detection tasks.
- For classification with constrained labels, preserve output
labels.
- For coding: keep code and exact literal-string transformations unchanged.
- If output is already in a fixed language required by the task, usually do not translate it.
- Proper names/titles may need to remain unchanged depending
on task context.
- Example: if task is translate English into French and the output is
French text, do not translate that output.
- Example: if input contains a film title or named entity that should remain original, avoid translating it.
- IMPORTANT: reason is only needed when decision is false.
- If decision is true, omit reason or leave it empty.
- Reasons must be very brief (max 8 words).
- If unsure, set should_translate=true.

Return ONLY valid JSON with exactly this schema:
{
  "instruction": {
    "all_languages": {"should_translate": true},
    "per_language": {}
  },
  "input": {
    "all_languages": {"should_translate": true},
    "per_language": {}
  },
  "output": {
    "all_languages": {"should_translate": true},
    "per_language": {}
  }
}
- Use language tags exactly from target list.
- all_languages is the default for every target language.
- per_language contains only exceptions.
No markdown. No extra keys.
\end{lstlisting}

\subsubsection{Task-Level Localisation}

\medskip
\noindent\textbf{System prompt}\par
\nobreak
\begin{lstlisting}[style=prompt]
You localise instruction-following dataset samples.
Return one valid JSON object only.
Do not use markdown code fences.
\end{lstlisting}

\medskip
\noindent\textbf{User prompt template}\par
\nobreak
\begin{lstlisting}[style=prompt]
You are creating a localised counterpart of one English Alpaca-style sample that cannot be directly translated.

Target language: {target_lang_name} ({target_lang})

Original sample metadata:
primary_task: {primary_task}
primary_domain: {primary_domain}
secondary_tasks: {secondary_tasks}
secondary_domains: {secondary_domains}

Original English sample:
instruction: {instruction}
input: {input}
output: {output}

Localisation rules:
- Write the final instruction, input, and output in {target_lang_name}.
- Make the localised sample as semantically close to the original as the target language allows.
- Treat the result as a near-parallel sample: preserve the same meaning, intent, task type, difficulty, constraints, and expected answer style whenever possible.
- The first priority is that the sample makes sense in the target language and that the task logic is correct.
- The second priority is semantic closeness to the original so the pair can reasonably be treated as parallel data.
- Do not do literal machine translation when it would break the task logic or sound unnatural in the target language.
- Adapt grammar, orthography, syntax, vocabulary, morphology, punctuation, and idioms to the target language.
- For spelling, grammar, tense, voice, punctuation, typos, word-count, rhyme, or wordplay tasks, create a target-language equivalent with target-language examples and a correct target-language answer.
- If the original sample uses an English-specific sentence, word, alphabet, proverb, idiom, or error, replace it with a natural equivalent in the target language.
- When replacing English-specific material, keep the replacement as close as possible in topic, tone, length, complexity, and answer structure.
- Preserve all functional information from the instruction/input fields, including separate lines, criteria, labels, options, constraints, and examples.
- Keep code, URLs, numbers, formal symbols, and named entities unchanged unless they are part of the language-specific exercise.
- Keep empty input empty when the localised task does not need an input field.
- The localised output must correctly answer the localised instruction/input, not the English source.
- Do not mention that the sample was translated or localised.
- Do not add explanations outside the requested output field.

Return ONLY valid JSON with exactly this schema:
{
  "instruction": "...",
  "input": "...",
  "output": "..."
}

No markdown. No extra keys. Do not wrap JSON in code fences.
Escape internal double quotes and newline characters correctly.
\end{lstlisting}

\subsubsection{Consistency Validation}
\vspace{-0.15em}
\medskip
\noindent\textbf{System prompt}\par
\nobreak
\begin{lstlisting}[style=prompt]
You validate multilingual instruction-following samples with strict language-specific coherence checks. Return JSON only.
\end{lstlisting}
\vspace{-0.15em}
\medskip
\noindent\textbf{User prompt template}\par
\nobreak
\begin{lstlisting}[style=prompt]
Task:
Validate one sample in language: {language_name}.

Input sample:
primary_task: {primary_task}
instruction: {instruction}
input: {input}
output: {output}

Coherence criteria (language-specific):
- Judge coherence from the actual text in {language_name}, not from the abstract task idea.
- Set is_coherent=false if wording in {language_name} makes the instruction/input/output semantically wrong, contradictory, or unclear.
- Set is_coherent=false if there is severe grammar/orthography/syntax degradation that changes or blocks meaning in {language_name}.
- Set is_coherent=false if mistranslation/literal calque causes wrong meaning in {language_name}.
- Set is_coherent=false if language mixing or wrong-language text breaks task understanding in {language_name}.
- If errors are minor and meaning remains clear and logically aligned, is_coherent can be true.

Language lock (mandatory):
- Final corrected instruction/input/output must be in {language_name}.
- Exception: keep spans in another language only when the task explicitly requires them (for example, translation tasks or quoted source text that must remain in the source language).
- Keep named entities, code, URLs, numbers, and formal symbols unchanged unless clearly malformed.
- Outside those required spans, avoid language mixing and normalise wording to {language_name}.

Task and constraint preservation:
- Treat explicit instruction constraints as part of the task meaning, including format, number of items, casing, punctuation restrictions, required/forbidden words, exact phrases, and requested response language.
- Set is_coherent=false if the instruction/input/output is internally inconsistent with an explicit instruction constraint, or if the output clearly violates one.

Return ONLY JSON with this schema:
{
  "is_coherent": true,
  "issues": "",
  "changes": {}
}

Rules:
- The issues field must contain a very short reason only when not coherent.
- If coherent, issues must be an empty string.
- Always correct grammar, orthography, punctuation, and syntax when needed.
- Put edits only in changes. The changes object must include ONLY modified fields among instruction/input/output as optional keys.
- If nothing needs editing, return changes as {}.
- If is_coherent=false, changes must repair the sample so it becomes coherent.
- If is_coherent=true, changes can still contain language fixes.
- Keep text in the dataset language, except required cross-language spans.
- Preserve original intent, constraints, factual content, and task type whenever possible.
- Do not add new requirements not implied by the sample.
- Keep URLs, code, numbers, and named entities unless clearly malformed.
- Keep empty input empty when appropriate.
- Do not use markdown fences. Do not return multiple JSON objects. Do not explain.
\end{lstlisting}

\end{document}